\documentclass{article} %
\usepackage{iclr2027_conference,times}

\usepackage{amsmath,amssymb,amsthm}
\usepackage{xcolor}
\usepackage{booktabs}
\usepackage{graphicx}

\usepackage{amsmath,amsfonts,bm}

\def\eqref#1{equation~\ref{#1}}

\def\1{\bm{1}}

\def\va{{\bm{a}}}
\def\vb{{\bm{b}}}

\def\ve{{\bm{e}}}

\def\vg{{\bm{g}}}

\def\vm{{\bm{m}}}

\def\vr{{\bm{r}}}

\def\vu{{\bm{u}}}
\def\vv{{\bm{v}}}
\def\vw{{\bm{w}}}

\def\vy{{\bm{y}}}

\def\mJ{{\bm{J}}}

\def\mW{{\bm{W}}}

\DeclareMathAlphabet{\mathsfit}{\encodingdefault}{\sfdefault}{m}{sl}
\SetMathAlphabet{\mathsfit}{bold}{\encodingdefault}{\sfdefault}{bx}{n}

\newcommand{\R}{\mathbb{R}}

\usepackage{hyperref}
\usepackage{url}
\hypersetup{pdftitle={Grokking Can Be Steered: Selecting Desirable Solutions By Choice Of Learning Rule},pdfauthor={Pascal Sager, Benjamin Grewe, Pau Vilimelis Aceituno, Thilo Stadelmann}}

\newtheorem{proposition}{Proposition}

\title{Grokking Can Be Steered: Selecting Desirable Solutions By Choice Of Learning Rule}%

\author{\makebox[0.45\textwidth][l]{Pascal Sager} \\
Centre for Artificial Intelligence, ZHAW \\
Institute of Neuroinformatics \\
ETH Z\"urich and University of Z\"urich \\
ETH AI Center, Z\"urich, Switzerland \\
\texttt{sage@zhaw.ch} \\
\And \makebox[0.45\textwidth][l]{Benjamin Grewe} \\
Institute of Neuroinformatics \\
ETH Z\"urich and University of Z\"urich \\
ETH AI Center, Z\"urich, Switzerland \\
\AND \makebox[0.45\textwidth][l]{Pau Vilimelis Aceituno\thanks{Shared last authorship.}} \\
Institute of Neuroinformatics \\
ETH Z\"urich and University of Z\"urich \\
\And \makebox[0.45\textwidth][l]{Thilo Stadelmann\footnotemark[1]} \\
Centre for Artificial Intelligence, ZHAW \\
Winterthur, Switzerland \\
ECLT, Venice, Italy }

\iclrfinalcopy
\begin{document}

\maketitle
\lhead{Preprint}

\begin{abstract}
During grokking, a network transitions from memorization to generalization, yet what shapes the neural implementation (circuit) of the generalizing solution and specifically the number of participating neurons (its size) remains unclear. We find that the learning rule, the combination of credit assignment algorithm and optimizer, controls this: On modular addition, every learning rule we study that groks recovers the same Fourier-based algorithm, yet implemented by circuits that differ in size. Why? A neuron stays in use when its credit is large relative to the non-gradient term of its own update, and this single condition unifies three routes to unused neurons: the coupled weight decay, Adam's $\epsilon$ and the decay of a trainable per-neuron gain. Equalizing the magnitude of credit assigned to each neuron, or scaling each neuron's weight decay by its credit, expands the effective circuit size, measured from the outgoing weights, from $60\%$ to $99\%$ of the layer's neurons at similar training and test accuracy. We vary which neurons share one equalization factor, from the whole layer to individual neurons, and find that the circuit stays concentrated for groups of eight or more and fully distributes only when every neuron has its own factor. Credit magnitudes are usually set by the built-in normalization of standard optimizers: Nero, which normalizes per neuron, distributes the computation, whereas Adam and seven others, including RMSProp, Muon, and Shampoo, use other normalization schemes and retain small, concentrated circuits. Distributed circuits tolerate neuron deletion better, while concentrated circuits can be pruned further and transfer about twice as fast to a task that recombines the same features, largely by recruiting their unused neurons. Learning rule choice thus steers circuit size, trading robustness against compressibility and transfer speed without changing the learned algorithm.
\end{abstract}

\section{Introduction}
A neural network can reach nearly perfect training accuracy while test accuracy remains near zero for many epochs before improving abruptly. This delayed generalization is called \emph{grokking} \citep{power2022grokking}. One can distinguish two aspects of a grokked solution: the computational procedure used to compute the function, which is called its \emph{algorithm}, and how that procedure is implemented across neurons, which is called its \emph{circuit}. Modular addition is a well-understood example of grokking and a central task in mechanistic interpretability of small networks. Its algorithm and circuit can both be analyzed at the neuron level: the network represents each operand as an angle on a circle at a few frequencies and adds the angles, a periodic computation carried by the neurons tuned to those frequencies \citep{nanda2023progress,gromov2023grokking,chughtai2023toy}. When we say that two networks learn the same algorithm, we mean that they share this Fourier-based computation (Appendix~\ref{app:algo}).

\begin{figure}[t]
\begin{center}\includegraphics[width=\linewidth]{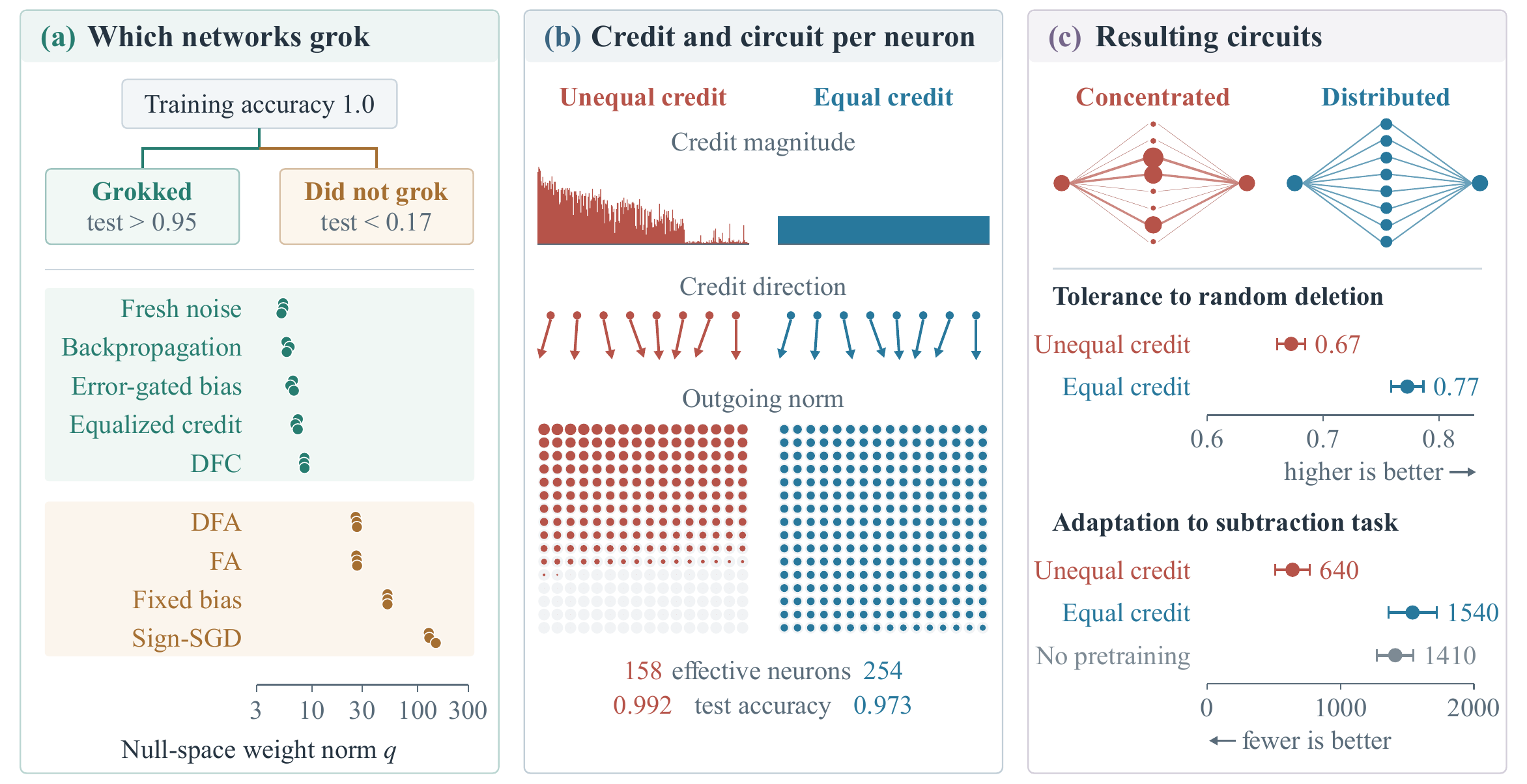}\end{center}
\caption{The learning rule selects whether a network groks, how many neurons implement the grokked algorithm, and the resulting robustness and adaptation. \textbf{(a)} All runs fit the training set, and only some then generalize to the held-out pairs. Each dot is one seed and shows the null-space weight norm $q$ at the end of training (Section~\ref{sec:grokrules}). \textbf{(b)} First hidden layer of two networks that share data and initialization. The credit of a neuron is the gradient of its incoming weights. Under backpropagation with Adam the credit is unequal (top row, relative to the layer mean and averaged over training), and neurons that receive little credit end with small outgoing weights. Credit equalization makes the magnitudes equal before the gradient enters the optimizer (Adam) and keeps every gradient's direction (middle row, schematic). Dots show the $256$ neurons in the order of the top row, with area proportional to the outgoing norm. Effective counts are for this seed, and the three-seed means are $154$ and $254$. \textbf{(c)} Tolerance is the normalized area under the curve of test accuracy against the fraction of randomly deleted neurons. Adaptation is the number of epochs to test accuracy $0.9$ on subtraction, starting from the addition network or from scratch (without pretraining). The network drawings are schematic, $\mu \pm \sigma$ are over ten seeds (Table~\ref{tab:adapt} and Appendix~\ref{app:deletion}).}
\label{fig:main}
\end{figure}

The concentrated (i.e., small) circuits reported for modular addition leave open whether the concentration is required by the task or produced by the training procedure. \citet{chughtai2023toy} describe the circuits learned for group composition in terms of group representations, but find that which representations a network learns, and in which order, varies across runs. Optimization research offers a potential explanation: the \emph{learning rule}, i.e., the combination of credit assignment algorithm and optimizer, \emph{chooses} among the many parameter settings that fit the training data \citep{gunasekar2018implicit,woodworth2020kernel}. Yet the networks in these circuit studies have all been trained with similar learning rules based on backpropagation. 

In this paper, we ask how the learning rule selects the circuit that implements a learned algorithm: In our experiments on modular addition, the same periodic computation appears in circuits ranging from concentrated to nearly fully distributed. Effective circuit size varies across learning rules but little across random seeds. We investigate which property of the learning rule controls concentration and what the resulting differences imply for robustness and adaptability (Figure~\ref{fig:main}). Our object is the circuit of the grokked solution. Modular arithmetic provides a setting in which its algorithm is known, and the redistribution also appears on sparse parity, which generalizes without a delay (Appendix~\ref{app:parity}).

Our central hypothesis is that circuit concentration depends on how the learning rule distributes credit across neurons, where \emph{credit} denotes the learning signal assigned to a neuron. Backpropagation with Adam assigns unequal credit, and the neurons that receive little of it end with small outgoing weights (Figure~\ref{fig:main}b). To isolate this effect, we modify backpropagation so that every neuron receives the same credit magnitude while its gradient direction is preserved, and we compare runs that share the data, initialization, loss, optimizer and hyperparameters. Equalizing credit increases the effective neuron count, computed from the outgoing weight norms, from approximately $154$ to $254$ of $256$ neurons in the first hidden layer while retaining the periodic computation. Training loss remains the same within a few percent, and test accuracy is about two percentage points lower. Applied after grokking, the same intervention redistributes the outgoing weights of an already learned circuit at unchanged test accuracy. A neuron stays in use when its credit is large relative to the non-gradient term of its update: the coupled weight decay in our reference, and Adam's $\epsilon$ or the decay of a normalization gain in other settings. This one condition accounts for three routes to unused neurons reported in separate lines of work \citep{yaguchi2018adam,mehta2019implicit,dufortlabbe2025maxwell,kolb2025dgating}, and rebalancing each neuron undoes all three.

Both parts of the learning rule shape the credit: the credit assignment algorithm computes each neuron's signal, and the optimizer rescales it. Rebalancing each neuron separately distributes the circuit, while groups of eight or more neurons that share one factor stay as concentrated as the reference. Among eight optimizers only Nero, which normalizes per neuron, produces a distributed circuit, and so does deep feedback control (DFC) \citep{meulemans2021dfc}, whose credit is nearly equal across neurons (Section~\ref{sec:law}).

Our contributions are the following. (i) Credit equalization, a controlled intervention on the learning rule that changes circuit concentration while the algorithm and the training fit stay the same and test accuracy stays within about two percentage points, including when applied after grokking. It shows that three routes to unused neurons, the coupled weight decay, Adam's $\epsilon$ and a decayed per-neuron gain, follow one per-neuron condition: the size of a neuron's credit relative to the non-gradient term of its update. (ii) A controlled comparison shows that normalizing each neuron separately produces nearly uniform outgoing-weight participation, whereas normalizing larger groups retains concentration. (iii) A trade-off: distributed circuits better tolerate neuron deletion, whereas concentrated circuits can be pruned further and adapt about twice as fast from addition to subtraction, a task that recombines the learned operand frequencies, largely by recruiting their nearly unused neurons. The circuit that pruning extracts for the same algorithm changes several-fold with the training configuration. The redistribution holds on modular subtraction and multiplication, permutation composition in $S_5$, sparse parity and a transformer with a wide MLP, and changes little where the reference circuit is already distributed (Appendix~\ref{app:scope}).

\section{Background and related work}
\label{sec:background}

\paragraph{Grokking and the assumed solution}
Grokking was first reported on modular arithmetic and other small symbolic tasks \citep{power2022grokking} and later on image classification \citep{liu2022omnigrok}. Existing research explains the emergence of generalizing computations through circuit efficiency, that is, larger logits at a given parameter norm, which weight decay rewards \citep{varma2023circuit}, through the weight norm at initialization \citep{liu2022omnigrok}, and through a transition from nearly fixed initial features to learned task-specific features \citep{kumar2024grokking}. \citet{nanda2023progress} track the Fourier circuit as it forms before test accuracy rises. A related theoretical analysis connects frequency selectivity with circuit size: after grokking, a one-hidden-layer network approaches solutions with single-frequency neurons and a computation distributed across many neurons \citep{morwani2024feature}.

Other studies report multiple solutions, but these differ in their algorithms rather than only in the number of units used. Changing the architecture, hyperparameters or seed can produce a different algorithm \citep{zhong2023clock}. A sparse generalizing subnetwork can also coexist with a dense memorizing subnetwork \citep{merrill2023tale}. Smaller initialization scales can concentrate a two-layer network into fewer effective neurons \citep{luo2021phase}, without identifying which circuit will form. Closest to our finding, \citet{manningcoe2025grokking} compare networks that grok with networks that generalize without delay and find that both learn the same features but encode them with different efficiency. Our comparison instead isolates credit allocation and measures how it distributes the circuit that implements the same algorithm.

\paragraph{What the learning rule selects}
The implicit-bias literature studies how optimization selects a solution among the many parameter settings that fit the training data. Such selection has been characterized for steepest descent under different norms \citep{gunasekar2018implicit} and for product parameterizations with small initialization \citep{woodworth2020kernel}. On grokking tasks, parameterization scale affects feature learning \citep{kumar2024grokking}, and the optimizer affects whether grokking occurs \citep{thilak2022slingshot}. With adaptive optimizers, an L2 penalty added to the gradient is rescaled together with it, so parameters with small gradients are regularized more strongly than under decoupled weight decay \citep{loshchilov2019decoupled}. We connect this selection to circuit concentration through a controlled intervention on credit allocation.

\paragraph{Unused units under adaptive optimizers}
Adam with an L2 penalty drives whole nodes and filters to zero \citep{yaguchi2018adam,mehta2019implicit}, Adam's $\epsilon$ lets units with tiny second-moment estimates die \citep{dufortlabbe2025maxwell}, decayed per-unit gains act as sparsity-inducing gates \citep{mehta2019implicit,kolb2025dgating}, and persistently small row-wise updates leave neurons dead under Muon in large language models \citep{dewulf2026aurora}. Under coupled L2 a neuron's equilibrium depends on its gradient norm \citep{kosson2024rotational}. These works treat each route separately, as a pathology or a tool for pruning. We show that three of them share one per-neuron condition, that rebalancing each neuron controls all three, and that the condition decides how many neurons implement a fixed algorithm.

\paragraph{Alternatives to backpropagation}
Several methods compute learning signals without using standard backpropagation. These include feedback alignment \citep{lillicrap2016feedback,nokland2016dfa}, predictive coding \citep{whittington2017pc}, equilibrium propagation \citep{scellier2017eqprop}, and deep feedback control (DFC) with its least control principle \citep{meulemans2021dfc,meulemans2022lcp}. They are usually evaluated by accuracy, which does not establish whether they learn the same computation or circuit as backpropagation. \citet{refinetti2021align} show that feedback alignment initially aligns with the gradient and later memorizes. \citet{mallinar2025agop} demonstrate grokking without backpropagation in recursive feature machines, a kernel method without a neuron-level circuit.

\section{Experimental setup}
\label{sec:setup}

\paragraph{Task and network.} We study modular arithmetic, a central task in grokking research (Section~\ref{sec:background}). The operands $a, b \in \mathbb{Z}_p = \{0, 1, \ldots, p-1\}$ are represented by concatenated one-hot vectors, and the target is $c = (a \circ b) \bmod p$ with $\circ \in \{+, -, \times\}$. Our primary task is addition modulo the prime $p = 31$. We use a fully connected ReLU network with two hidden layers of width $256$ and a linear readout; multilayer perceptrons of this kind are standard in grokking studies of modular arithmetic \citep{gromov2023grokking,morwani2024feature}. Unless stated otherwise, circuit measurements and neuron deletion concern the first hidden layer.

\paragraph{Training.} Our reference learning rule is backpropagation with Adam \citep{kingma2015adam}, cross-entropy loss and coupled weight decay $10^{-3}$, which adds the decay term to the gradient before Adam's moment estimates (L2 regularization); Section~\ref{sec:rules} shows that this choice matters. Runs that we compare with each other share the data split and the initialization. We hold out $40\%$ of the operand pairs, train full batch, and apply credit equalization to the first hidden layer; budgets are stated with each experiment (Appendix~\ref{app:repro}). We classify a run as grokking when training accuracy reaches $1.0$ and test accuracy first exceeds $0.95$ at least $500$ epochs later. Results are means over three seeds unless stated otherwise, with standard deviations in parentheses or as error bars.

\subsection{Learning rules}
\label{sec:ruleset}
We distinguish two parts of training. A \emph{credit assignment algorithm}, such as backpropagation, computes a learning signal for each parameter. We call the signal received by a hidden neuron its \emph{credit}. An \emph{optimizer}, such as SGD or Adam, converts these signals into weight updates. We use \emph{learning rule} for the combination of the two and vary each part in turn (Table~\ref{tab:ruleset}).

Credit equalization is the intervention we introduce. It rescales the magnitude of each neuron's incoming-weight gradient before the gradient enters Adam and leaves its direction unchanged (Section~\ref{sec:rules}). The optimizer comparison keeps backpropagation and replaces Adam with LARS, RMSProp, LAMB, Adafactor, Lion, Muon, Shampoo or Nero. DFC replaces the loss gradient by a controller-based signal. During training, a feedback controller drives the network output toward a target, its signal reaches each layer through the transposed Jacobian of the forward network, and each layer learns from the difference between its controlled and its feedforward activity \citep{meulemans2021dfc}. Section~\ref{sec:grokrules} asks which rules grok at all and adds six rules for that purpose. Feedback alignment and direct feedback alignment replace the transposed forward weights of the backward pass by fixed random matrices \citep{lillicrap2016feedback,nokland2016dfa}, and sign-SGD keeps only the sign of each gradient coordinate \citep{bernstein2018signsgd}. Three constructed rules add a component orthogonal to the gradient that is fresh at every step, fixed for the run, or fixed and fading with the training loss (Appendix~\ref{app:generalization}). All six signals are passed to Adam, and the rules that memorize do not enter the circuit comparisons.

\begin{table}[t]
\caption{Learning rules under test, grouped by how they transform the learning signal. Rules that enter only Section~\ref{sec:grokrules} are not listed. Settings are given in Appendix~\ref{app:repro}.}
\label{tab:ruleset}
\begin{center}
\footnotesize
\setlength{\tabcolsep}{4pt}
\begin{tabular}{@{}p{0.34\linewidth}p{0.63\linewidth}@{}}
\toprule
Learning rule & Transformation of the learning signal \\
\midrule
Backpropagation with Adam (reference) & per-parameter adaptive scaling \citep{kingma2015adam} \\
Credit equalization, then Adam & one positive factor per neuron, applied before Adam \\
\midrule
LARS & one factor per layer \citep{you2017lars} \\
LAMB & per-parameter scaling and a layerwise trust ratio \citep{you2020lamb} \\
RMSProp & per-parameter second-moment scaling \citep{tieleman2012rmsprop} \\
Lion & elementwise sign of a momentum signal \citep{chen2023lion} \\
Adafactor & factored second-moment scaling \citep{shazeer2018adafactor} \\
Muon & matrix orthogonalization \citep{jordan2024muon} \\
Shampoo & matrix preconditioning \citep{gupta2018shampoo} \\
Nero & one factor per neuron and weight constraints \citep{liu2021nero} \\
\midrule
DFC & controller-based credit, no gradient rescaling \citep{meulemans2021dfc} \\
\bottomrule
\end{tabular}
\end{center}
\end{table}

\section{Results: Credit allocation changes circuit concentration}
\label{sec:select}

\paragraph{Measuring circuit concentration.} We ask how a learned computation is spread across the neurons of a layer in the networks that grok. On modular addition, algorithm checks for all the tested learning rules that grok show the same periodic Fourier computation and not the alternative computation reported by \citet{zhong2023clock} (Appendix~\ref{app:algo}). The checks cover the input frequencies used, a trigonometric fit of the logits, and agreement across seeds and rules; for the reference and the equalized network, the fitted trigonometric function alone classifies every held-out pair correctly.

We measure circuit concentration from the norms of the weights that connect each hidden neuron to the next layer, since, at a given activation, a neuron with small outgoing weights has only a small effect on the output, whatever it computes. For the $n$ neurons in the measured layer, let $a_j = \|\mW^{\mathrm{out}}_{:,j}\|_2$ be the norm of the outgoing weights of neuron $j$. Their \emph{participation ratio} is
\begin{equation}
\mathrm{PR} \;=\; \frac{\big(\sum_j a_j\big)^2}{n \sum_j a_j^2} \;\in\; [1/n, 1],
\label{eq:pr}
\end{equation}
for a nonzero vector of norms. If all norms are equal, $\mathrm{PR} = 1$, and if only one is nonzero, $\mathrm{PR} = 1/n$ (Proposition~\ref{prop:pr}, and all propositions are stated and proved in Appendices~\ref{app:theory} and~\ref{app:generalization}). Applied to the outgoing norms, as here, we call it the \emph{circuit participation}, and $n \cdot \mathrm{PR}$ the \emph{effective neuron count} (a layer with effective count $154$ has the participation ratio of a layer in which $154$ neurons carry equal outgoing norms and the rest none). We call a neuron \emph{nearly unused} if its outgoing norm is below a tenth of the layer mean. Both are computed from the weights, and Section~\ref{sec:rules} confirms them with two measures tied to the computation. Applied to the norms of the learning signal that the neurons receive (Section~\ref{sec:law}), we call it the \emph{credit participation}.

Backpropagation with Adam produces a concentrated circuit with approximately $154$ effective neurons out of $256$ in the first hidden layer, and about $84$ of the $256$ neurons are nearly unused. In contrast, DFC groks with a delay of $520$ to $610$ epochs (Appendix~\ref{app:repro}) and produces a distributed circuit with approximately $235$ effective neurons at nearly perfect test accuracy.

\subsection{Equalizing credit across neurons}
\label{sec:rules}

Backpropagation and DFC differ in several respects, so we modify backpropagation itself and keep everything else fixed.

Let $\vg_j$ be the gradient of neuron $j$'s incoming weights. Its norm $\|\vg_j\|_2$ is the neuron's credit magnitude. The mean credit magnitude over the $n$ neurons of the layer is $\bar{g} = \frac{1}{n}\sum_{k=1}^{n} \|\vg_k\|_2$. The intervention moves each magnitude toward the mean with strength $\beta \in [0,1]$,
\begin{equation}
\widetilde{\vg}_j \;=\; \frac{\vg_j}{\|\vg_j\|_2}\Big( (1-\beta)\, \|\vg_j\|_2 + \beta\, \bar{g} \Big)
\quad \text{for } \|\vg_j\|_2 > 0,
\label{eq:knob}
\end{equation}
and leaves zero gradients at zero. At $\beta = 0$ the gradient is unchanged and training is ordinary backpropagation with Adam. At $\beta = 1$ all nonzero incoming-weight gradients have the same norm. We call this \emph{full credit equalization}. The direction of every neuron's incoming-weight gradient is preserved, so before it enters Adam the rescaled gradient is still a descent direction for the loss (Proposition~\ref{prop:descent}, Appendix~\ref{app:theory}). The rescaling acts only on the incoming-weight gradients of the first hidden layer. Circuit participation is measured from the outgoing weights of the same neurons, which are the incoming weights of the second hidden layer and are not rescaled.

\begin{figure}[t]
\begin{center}\includegraphics[width=\linewidth]{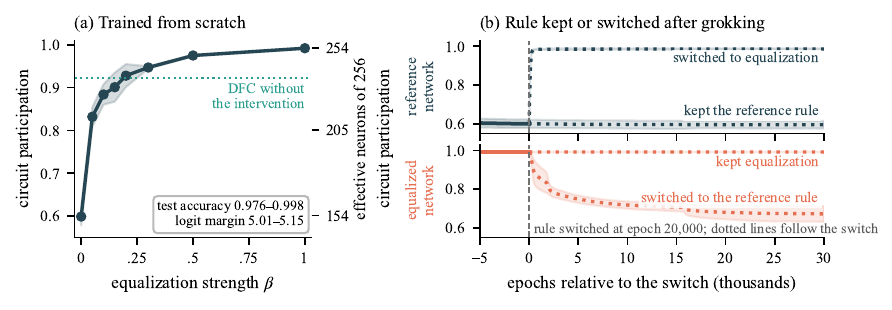}\end{center}
\caption{Credit allocation sets the circuit and reorganizes it after grokking (modular addition, $p=31$, three seeds, bands one standard deviation). \textbf{(a)} Circuit participation and effective neuron count of networks trained from scratch at equalization strength $\beta$. Training cross-entropy ($0.040$), margin and parameter norm stay close to the reference values. \textbf{(b)} A network grokked with the reference rule and one grokked with credit equalization, each copied at epoch $20{,}000$ and continued for $30{,}000$ epochs with its own rule or with the other rule. Test accuracy remains above $0.95$ throughout.}
\label{fig:steer}
\end{figure}

Increasing $\beta$ from $0$ to $1$ raises circuit participation from $0.60$ to $0.99$, that is, from approximately $154$ to $254$ effective neurons (Figure~\ref{fig:steer}a). From $\beta = 0.25$ on no neuron is nearly unused, yet participation keeps rising, from $0.93$ to $0.99$ (Appendix~\ref{app:adapt}). Training loss, margin (the mean difference between the correct logit and the largest incorrect logit) and parameter norm stay within a few percent of the reference values, and mean test accuracy decreases from $0.998$ to $0.976$. The gap widens with width: at width $2048$ the reference uses $332$ effective neurons and the equalized network $2046$ (Appendix~\ref{app:arch}). The circuit concentrates during the grokking delay. In the reference, participation is $0.997$ when the training set is first fitted, has fallen to between $0.59$ and $0.67$ when test accuracy reaches $0.95$, and changes by at most $0.05$ afterwards (five seeds, Appendix~\ref{app:timing}).

Two measures that use each neuron's activity or its effect on the output also increase. Over eight seeds, participation computed from outgoing norms weighted by mean absolute activation rises from $0.34$ to $0.81$. Participation computed from the absolute mean change in the correct-class logit after removing each neuron rises from $0.42$ to $0.80$ (Appendix~\ref{app:functional}). The activation-weighted measure is invariant to compensating rescalings of a ReLU neuron's weights.

We next ask whether the intervention can redistribute a circuit that has already been learned. We let the reference network grok, then switch to full credit equalization and continue training on the same task. Outgoing-weight participation rises from $0.60$ to above $0.95$ within $300$ epochs and then to $0.99$, while test accuracy remains above $0.98$ (Figure~\ref{fig:steer}b). The effect of the neurons on the output spreads as well, from ablation participation $0.40$ to $0.72$ over five seeds (Appendix~\ref{app:adapt}). A control that continues with the reference rule stays at $0.60$, and in separate runs both rules hold their participation over $200{,}000$ epochs (Appendix~\ref{app:long}).

Controls in Appendix~\ref{app:credit_controls} exclude alternative explanations. Constraining the incoming weight norms instead of the credit gives participation $0.66$, and a grid over learning rate and weight decay stays between $0.53$ and $0.63$. Dropout also distributes the circuit, but it changes the objective and raises margin and norm (Appendix~\ref{app:baselines}). In a simplified linear model under gradient descent, each unit grows in proportion to its size without equalization and by the same amount with it, so initial differences persist in the first case and fade in the second (Proposition~\ref{prop:select}, Appendix~\ref{app:theory}).

\paragraph{A per-neuron condition.} A neuron stays in use when its credit is large relative to the non-gradient term of its own update. Adam is nearly invariant to a constant rescaling of its input, so rescaling a neuron's gradient changes its size relative to the other terms of its update, in our reference the coupled decay term added to the gradient before Adam. Rescaling only this term per neuron, with the gradient unchanged, reproduces the redistribution, and equalizing the gradient while holding each neuron's ratio of gradient to decay leaves the circuit concentrated, on the activation-weighted and ablation measures as well (Table~\ref{tab:pathways}, Appendix~\ref{app:decay}). The same condition holds for two other terms that do not scale with the gradient: Adam's $\epsilon$, which dominates the denominator of its update when AdamW decays a transformer's normalization parameters (Appendix~\ref{app:transformer}), and the decay of a trainable per-neuron gain and of the outgoing weights (Appendix~\ref{app:normlayers}). Lion and SGD with coupled decay follow it as well (Appendix~\ref{app:baselines}). Where no non-gradient term dominates, as under AdamW without decay on normalization parameters, the reference is already broadly distributed, with participation $0.91$.

\begin{table}[t]
\caption{Rescaling only the non-gradient term of each neuron's update reproduces credit equalization in all three cases. Equalizing the gradients while holding each neuron's ratio of credit to that term fixed keeps the circuit concentrated for the decay and the gain and removes about half of the effect for $\epsilon$ (circuit participation, five seeds per cell). For the gain, equalization and rescaling act on the gain and the outgoing weights as well as the incoming rows; equalizing the incoming rows alone gives $0.364$. Update equations and details in Appendices~\ref{app:decay}, \ref{app:transformer} and~\ref{app:normlayers}.}
\label{tab:pathways}
\begin{center}
\footnotesize
\setlength{\tabcolsep}{4pt}
\begin{tabular}{llcccc}
\toprule
Non-gradient term & Network & Reference & Equalized & Term rescaled & Ratio held \\
\midrule
Coupled decay & MLP, Adam with L2 & $0.606$ & $0.993$ & $0.993$ & $0.574$ \\
Adam's $\epsilon$ & Transformer, AdamW & $0.609$ & $0.994$ & $0.981$ & $0.818$ \\
Decay of the gain & MLP, layer normalization & $0.366$ & $0.996$ & $0.997$ & $0.343$ \\
\bottomrule
\end{tabular}
\end{center}
\end{table}

\subsection{Which parameters are normalized together?}
\label{sec:law}

Full credit equalization is itself a normalization of the gradient with one factor per neuron. Optimizers normalize the learning signal at other scales: Adam normalizes each weight separately, LARS rescales a whole layer by one factor, and Nero normalizes the gradient of the weights into each neuron. Before comparing optimizers, we ask in a controlled setting whether normalizing several neurons together has the same effect as normalizing each one alone. We divide the $n$ neurons of the first hidden layer into groups of $m$ neurons and rescale all gradients in a group by one common factor, chosen so that every group has the same mean credit,
\begin{equation}
\widetilde{\vg}_j \;=\; \vg_j\, \frac{\bar{g}}{\bar{g}_{B(j)}},
\qquad
\bar{g}_{B(j)} \;=\; \frac{1}{m}\sum_{k \in B(j)} \|\vg_k\|_2,
\label{eq:grain}
\end{equation}
where $B(j)$ is the group containing neuron $j$ and $\bar{g}$ is the layer mean from Equation~\ref{eq:knob}. With $m = 1$ every neuron is rescaled separately and the rule is full credit equalization ($\beta = 1$). With $m = n$ the factor equals one and the gradient is unchanged.

\begin{figure}[tb]
\begin{center}\includegraphics[width=\linewidth]{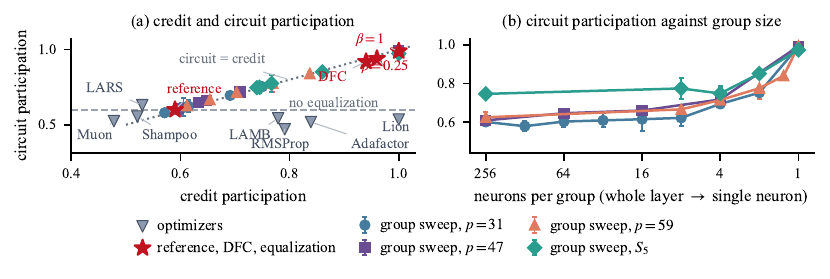}\end{center}
\caption{Circuit participation under different transformations of the learning signal. \textbf{(a)} Circuit participation against credit participation for the reference, DFC, credit equalization at $\beta = 0.25$ and $\beta = 1$, the group-size sweep on four tasks and seven optimizers. Credit participation is measured on the signal listed in Table~\ref{tab:creditconv}. \textbf{(b)} Circuit participation against the number of neurons that share one normalization factor, on modular addition at three moduli and on $S_5$. Error bars show one standard deviation across three seeds.}
\label{fig:law}
\end{figure}

For groups of eight or more neurons, circuit participation stays within about $0.02$ of the value $0.60$ of the reference, backpropagation with Adam. It rises to $0.70$ for groups of four, $0.75$ for pairs and $0.99$ for single neurons (Figure~\ref{fig:law}b). Most of the redistribution occurs only when each neuron receives its own factor. This equalizes the mean credit between groups but keeps the ratios inside a group: a neuron that received twice the credit of its neighbor still does after the rescaling. In a simplified model, a linear network in which every unit has one incoming and one outgoing scalar weight, these within-group differences persist throughout training and keep the circuit concentrated. Only single-neuron groups leave none to preserve (Proposition~\ref{prop:grain} in Appendix~\ref{app:theory}, sweep details in Appendix~\ref{app:grain}).

\paragraph{Comparison with optimizers.} If per-neuron normalization is what distributes the circuit, optimizers should sort accordingly. We train with the eight optimizers of Table~\ref{tab:ruleset} that replace Adam. LARS, RMSProp, LAMB, Adafactor, Lion, Muon and Shampoo normalize groups other than a single neuron or change update directions, and all seven produce concentrated circuits, with participation between $0.470$ and $0.613$. All seven run with the coupled weight decay of the reference. Nero, which uses no weight decay, produces at its default learning rate of $10^{-2}$ a distributed circuit with participation $0.986$ at test accuracy $0.985$ (Table~\ref{tab:zoo}). Nero, which normalizes credit per neuron, and DFC, whose credit is nearly equal (credit participation $0.94$), are cases of equal credit that arise without our intervention, and both produce distributed circuits.

For the reference, DFC and the equalization sweeps, credit participation lies close to circuit participation (Figure~\ref{fig:law}a). The condition also explains why equal update norms do not suffice. Lion gives every neuron an update of the same norm, and Adam-mini \citep{zhang2025adammini} normalizes each neuron's update with one second-moment estimate, yet both keep the circuit concentrated (Adam-mini $0.586$ against $0.594$ for Adam, Appendix~\ref{app:grain}). Both normalize the coupled decay together with the gradient, so each neuron's ratio of credit to decay is unchanged.

\subsection{Which learning rules grok}
\label{sec:grokrules}
On modular addition, backpropagation with the optimizers in Table~\ref{tab:ruleset}, credit equalization, DFC and the two constructed rules with a fresh or a fading orthogonal component grok, whereas feedback alignment, direct feedback alignment, sign-SGD and the rule with a fixed orthogonal component fit the training set and stay near zero test accuracy over $15{,}000$ epochs. This split matches the weight-decay account of grokking \citep{nanda2023progress,varma2023circuit}. Once the training data are fitted, weight decay removes weight along the directions in which small parameter changes leave the training outputs unchanged, and a learning signal that is not a loss gradient can hold weight there. The norm $q$ of the weight in these directions (Appendix~\ref{app:generalization}) separates the two outcomes at the end of training without overlap across nine configurations with three seeds each (Figure~\ref{fig:main}a). It shrinks to means between $5.3$ and $8.5$ in the rules that grok and stays between $26$ and $135$ in the rules that memorize. Within a rule, $q$ does not predict the outcome (Appendix~\ref{app:generalization}).

\section{Relevance: Distributed circuits tolerate neuron deletion, concentrated circuits keep spare capacity}
\label{sec:fragile}

Throughout this section the concentrated circuit is the reference ($\beta = 0$) and the distributed circuit is the fully equalized network ($\beta = 1$), both trained on modular addition at $p = 31$.

\subsection{Tolerance to neuron deletion}
A computation spread over many neurons should depend less on any single one of them \citep{georgopoulos1986population}, and networks that generalize well are known to rely less on single units \citep{morcos2018importance}. In an additive model, the relative fluctuation of the output under random deletion falls with the effective neuron count (Proposition~\ref{prop:fluct}, Appendix~\ref{app:theory}). We test this with three deletion procedures, all applied to the first hidden layer of the trained network without retraining. Random deletion removes a random subset of the neurons. Norm-ranked deletion removes neurons in decreasing order of outgoing norm, so the neurons a concentrated circuit relies on go first. Greedy deletion removes, at each step, the neuron whose removal lowers test accuracy the most. For random and norm-ranked deletion we report the \emph{neuron-deletion tolerance}, or \emph{deletion tolerance} for short, the normalized area under the curve of test accuracy against the deleted fraction from $0$ to $0.8$. For greedy deletion we report the number of neurons removed before test accuracy falls below $0.5$ (Appendix~\ref{app:deletion}).

\begin{table}[t]
\caption{Deletion tolerance and adaptation of the concentrated and the distributed circuit on modular addition ($p=31$). Means over three seeds for deletion, five for pruning to test accuracy $0.9$ (Appendix~\ref{app:deletion}), ten for subtraction and five for multiplication, with standard deviations in parentheses (Table~\ref{tab:adapt}).}
\label{tab:consequences}
\begin{center}
\footnotesize
\begin{tabular}{lccc}
\toprule
 & Concentrated ($\beta=0$) & Distributed ($\beta=1$) & From scratch \\
\midrule
Random deletion tolerance & $0.674$ & $0.770$ & --- \\
Norm-ranked deletion tolerance & $0.332$ & $0.581$ & --- \\
Greedy deletion, neurons removed & $23$ & $61$ & --- \\
Magnitude pruning, fraction removed & $0.62$ & $0.20$ & --- \\
Greedy pruning, neurons removed & $191$ & $160$ & --- \\
Adaptation to subtraction, epochs & $640\,(128)$ & $1540\,(180)$ & $1410\,(137)$ \\
Adaptation to multiplication, epochs & $1800\,(141)$ & $1940\,(136)$ & $1200\,(63)$ \\
\bottomrule
\end{tabular}
\end{center}
\end{table}

The distributed circuit is more tolerant under all three procedures (Table~\ref{tab:consequences}). The advantage is largest under norm-ranked deletion, as expected if the concentrated circuit rests on few high-norm neurons, and it persists under greedy deletion. In this network, the tolerance follows the distribution of outgoing norms rather than the rule that produced it: an outgoing-norm constraint that holds participation at $0.99$ without changing the gradients gives the same random-deletion tolerance as equalization (Appendix~\ref{app:suff}). Nero reaches similar deletion tolerances without the intervention (Appendix~\ref{app:zoo}).

Removing neurons in the opposite order, smallest outgoing norm first as in magnitude pruning, favors the concentrated circuit. It keeps test accuracy above $0.9$ until $62\%$ of its neurons are removed, against $20\%$ for the distributed circuit, because its nearly unused neurons can go at no cost. Greedy pruning gives the same order (Table~\ref{tab:consequences}). The circuit that pruning extracts for the same algorithm also depends on training. At width $2048$, where both start at test accuracy $0.998$, greedy pruning to $0.95$ keeps $33$ to $98$ neurons of the reference and $138$ to $229$ under equalization, and magnitude pruning keeps $143$ against $1692$ on average. A decayed per-neuron gain lowers the greedy circuit to $14$ to $22$ neurons, at least $7$ times fewer than under equalization in each seed (Appendix~\ref{app:minimal}).

\subsection{Adaptation to a new operation}
We take networks that have grokked addition at each $\beta$ and fine-tune the whole network on a new operation with backpropagation with Adam. The \emph{adaptation time} is the number of fine-tuning epochs until test accuracy on the new operation reaches $0.9$. We compare it with training a new network from scratch.

We first transfer to subtraction, which uses the same Fourier frequencies as addition and differs in how they are combined. Training from scratch takes $1410$ epochs. Starting from the concentrated addition circuit takes $640$ epochs, so the pretrained circuit helps. Starting from the distributed circuit takes $1540$ epochs, no faster than from scratch, and the adaptation time increases monotonically with $\beta$ in between (Appendix~\ref{app:adapt}). The concentrated circuit adapts about $2.4$ times as fast as the distributed one. The ordering holds at one third and at three times the fine-tuning learning rate, in the reverse transfer from subtraction to addition, when an outgoing-norm constraint sets the participation of the source network, and when both networks are fine-tuned with credit equalization instead of the reference rule (Appendices~\ref{app:suff} and~\ref{app:adapt}). The difference also appears when both continue from the same grokked checkpoint, with $1052$ against $608$ epochs, slower in all five seed pairs.

Much of the advantage comes from the nearly unused neurons. With them removed from the concentrated network at unchanged test accuracy, adaptation to subtraction takes $1204$ instead of $652$ epochs, against $1480$ for the distributed network, so they account for about two thirds of the difference. On $(b - a) \bmod p$ they account for about half (five seeds, Appendix~\ref{app:adapt}). Networks distributed by per-neuron decay alone adapt as slowly as equalized ones, so across the tested interventions adaptation follows the circuit, not the rule that produced it. Targets that keep the hidden features of addition and change only the readout, $(a + b + 7) \bmod p$ and a fixed permutation of $(a + b) \bmod p$, are learned $2.3$ to $6.6$ times faster than from scratch, with no consistent advantage for either circuit.

We repeat the experiment with multiplication, which does not share the Fourier frequency representation of addition. Every pretrained network takes $1800$ to $2020$ epochs, compared with $1200$ from scratch, so pretraining on addition does not help at any $\beta$.

\section{Discussion and conclusions}

Circuit concentration is a consequence of the training dynamics and not of the task alone. The concentrated circuits found under backpropagation with Adam are one of several ways a grokked solution can be implemented. A neuron stops contributing when its credit is small relative to the non-gradient term of its update, whether that term is the coupled decay, Adam's $\epsilon$ or the decay of a gain, and learning rules that rebalance each neuron keep it in use. The condition also governs MLP units in a $30$M-parameter language model trained with Adam and L2 regularization, where balancing each unit's decay leaves $1.3\%$ instead of $24\%$ nearly unused at lower validation loss, and AdamW leaves $0.9\%$, as the condition predicts (Appendix~\ref{app:lm}).

For mechanistic interpretability this means that the size of a circuit found in a trained network, including the circuit that pruning extracts, reflects the learning rule and how weight decay enters it, as well as the task. A sparse circuit read out of a network trained with Adam is a property of that training run, and comparing learning rules separates what the task requires from what the optimizer selected.

A distributed circuit loses less when units are removed, as under a hardware fault, while pruning and fine-tuning profit from the nearly unused neurons of a concentrated circuit, which can be removed at no cost or recruited by a related task that needs new hidden features. On modular addition, the choice of learning rule trades robustness to neuron loss against compressibility and spare capacity, without changing the learned algorithm.

\paragraph{Limitations.} The theory isolates the distribution mechanism in a simplified linear model without adaptive normalization and covers rescalings that preserve update directions. The results on nonlinear networks and on optimizers are empirical. Under AdamW without decay on normalization parameters the remaining effect is small (participation $0.91$ to $0.96$) and not explained by the condition, and for Adam's $\epsilon$ holding the per-neuron ratio fixed removes about half of the effect. The algorithm checks cover modular addition only. The adaptation advantage appears for targets that need new hidden combinations of the operand frequencies and not for targets that change only the readout. The robustness advantage concerns the deletion of neurons.

\subsection*{AI use statement}

We used generative AI tools to assist with experiment and analysis code, to edit text written by the authors, to check related-work coverage, and to check the derivations and proofs in the appendix. The checks of the derivations and proofs did not lead to changes. The authors developed the research ideas, experimental design, derivations, interpretations, and conclusions. We reviewed all AI-assisted work, read the code and tested it against its results, and checked suggested references and formulations. We take responsibility for the final text, claims and artifacts, including material developed with generative AI assistance.

\subsection*{Reproducibility statement}

Section~\ref{sec:setup} and Appendix~\ref{app:repro} specify the tasks, networks, learning rules and training settings. The code for all experiments, tables and figures will be published on GitHub after acceptance.

\bibliography{iclr2027_conference}
\bibliographystyle{iclr2027_conference}

\appendix

\section{Experimental details}
\label{app:repro}

\paragraph{Statistical reporting.} Results are means across three random seeds, with standard deviations in parentheses, unless stated otherwise. Conditions that are compared with each other use the same seeds and data splits. Independent experiments have their own reference runs, so reference values vary slightly between tables. For comparisons between credit and circuit participation we take the absolute difference within each seed and then average.

\paragraph{Networks and training.} Training is full batch. Hidden weights use Kaiming normal initialization \citep{he2015delving}. Adam uses learning rate $10^{-3}$, $\beta_1 = 0.9$, $\beta_2 = 0.999$, $\epsilon = 10^{-8}$ and coupled weight decay, in which the decay term is added to the gradient before the moment estimates; biases are trained and decayed like the weights. For the learning signals passed to Adam, the gradient norm is clipped to $1.0$ after any rescaling and before the optimizer step, and Adam adds the decay term to the clipped gradient. The optimizers compared in Section~\ref{sec:law} apply no clipping. In the multilayer perceptrons, credit equalization is applied to the incoming-weight gradients of the first hidden layer (bias gradients are not rescaled), and circuit participation and neuron deletion are measured on the outgoing weights of the same neurons. The transformer experiments apply it to every linear map in the canonical configuration and to the incoming weights of the MLP in the wide configuration (Appendix~\ref{app:transformer}). Seeds start at $0$, so three-seed experiments use seeds $0$ to $2$. Each run uses one GPU. Runs continue past grokking until participation has stabilized, which takes $12{,}000$ to $70{,}000$ epochs depending on task, modulus and architecture; compared conditions use the same budget, and the experiments with other budgets state them.

\paragraph{Modular-arithmetic data.} We hold out $40\%$ of the $p^2$ operand pairs at random and train on the rest, with the same split procedure for all three operations. Multiplication includes the pairs with a zero operand, identically across rules. We test $p \in \{31, 47, 59\}$ and use $p = 31$ unless stated otherwise.

\paragraph{Optimizer settings.} The comparison of Section~\ref{sec:law} runs every optimizer at the reference learning rate $10^{-3}$ and weight decay $10^{-3}$ for $20{,}000$ epochs. LARS uses fifty times this learning rate. Nero uses its default learning rate $10^{-2}$ and no weight decay, since its norm constraint fixes the weight scale; it normalizes each neuron's gradient by a running estimate of its norm and holds each neuron's incoming-weight norm fixed after its first update. Lion adds weight decay to the gradient before the signed momentum update. Muon uses five cubic Newton--Schulz iterations, ordinary momentum and coupled weight decay on matrices. LARS uses a layerwise norm ratio without momentum. LAMB uses uncorrected moments and caps the trust ratio at $10$. Adafactor uses fixed-rate factored second moments without update clipping, Shampoo uses exponential second moments, and RMSProp uses second-moment decay $0.99$ without momentum. Evaluated every $50$ epochs over three seeds, every optimizer fits the training set first and exceeds test accuracy $0.95$ at least $500$ epochs later, so all of them grok under the criterion of Section~\ref{sec:setup}: the delays are $500$ to $650$ epochs for LARS, $650$ to $700$ for Lion, $1200$ to $1350$ for RMSProp and Adafactor, $2150$ to $2400$ for LAMB, $7450$ to $7800$ for Muon, $12{,}200$ to $13{,}100$ for Shampoo and $4450$ to $7000$ for Nero, against $1300$ to $1550$ for the reference.

\begin{figure}[!ht]
\begin{center}\includegraphics[width=0.5\linewidth]{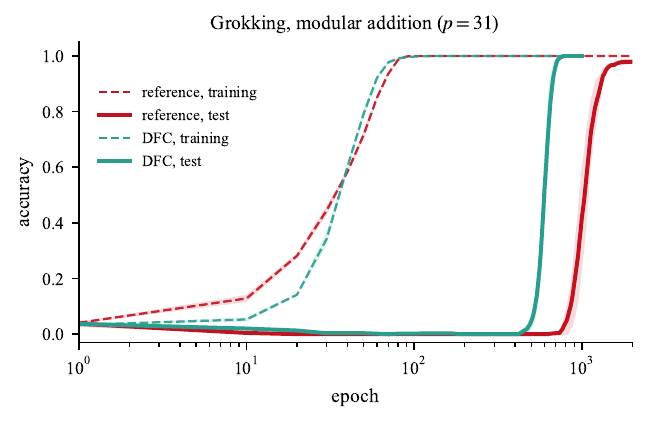}\end{center}
\caption{Training (dashed) and test (solid) accuracy for the reference and DFC on modular addition ($p=31$), three seeds with shared initializations, evaluated every $10$ epochs; bands show one standard deviation across seeds, and the DFC runs end at epoch $1000$. Both fit the training set first and generalize hundreds of epochs later.}
\label{fig:grok}
\end{figure}

\paragraph{Deep feedback control.} DFC uses a controller during training that moves the network output toward a target. The output $\hat{\vy}$ is the linear readout, that is, the logits, and the label $\vy$ is one-hot. The output error is $\ve = \vy^* - \hat{\vy}$, where the target $\vy^* = \hat{\vy} + 2\kappa(\vy - \hat{\vy})$ lies between the current output and the label at a distance set by the control strength $\kappa$; it is one step on the squared error to the label, as in the reference implementation. A leaky proportional-integral controller turns the error into a control signal $\vu$, which is fed back into every layer $i$ through the transpose of the Jacobian $\mJ_i$ from that layer to the output:
\begin{align}
\vu_{\mathrm{int}} &\leftarrow \vu_{\mathrm{int}} + \mathrm{dt}\,(\ve - \alpha \vu),
\qquad \vu = \vu_{\mathrm{int}} + k_p \ve, \label{eq:controller}\\
\vv_i &\leftarrow \vv_i + \tau ( \mW_i \vr_{i-1} + \vb_i + \mJ_i^\top \vu - \vv_i ),
\qquad \vr_i = \phi(\vv_i). \label{eq:dynamics}
\end{align}
We use $\mathrm{dt} = 0.02$, $\tau = \mathrm{dt}/0.2 = 0.1$, where $0.2$ is the ratio of the neuronal to the controller time constant, $k_p = 2.0$ and $\alpha = 1.7\times10^{-3}$, allow at most $500$ controller iterations, and stop when the change in the control signal falls below $10^{-4}$. The control strength is $\kappa = 0.3$ for modular arithmetic and $0.5$ for parity. The modular-arithmetic setting was selected on a grid over weight decay $\{10^{-3}, 3\times10^{-3}, 10^{-2}\}$ and $\kappa \in \{0.1, 0.2, 0.3\}$. At equilibrium each layer applies the local update
\begin{equation}
\Delta \mW_i \;\propto\; (\vr_i - \vr_i^{\mathrm{ff}})\, \vr_{i-1}^\top,
\label{eq:update}
\end{equation}
the difference between the controlled activity $\vr_i$ and the feedforward activity $\vr_i^{\mathrm{ff}}$, multiplied by the presynaptic activity. Its negative, $-2\,(\vr_i - \vr_i^{\mathrm{ff}})\,\vr_{i-1}^\top$ averaged over the batch, is written into the gradient slot, so that Adam's descent step moves the weights along $\Delta \mW_i$; it is clipped and passed to Adam like the other learning signals in this paper. Feedback through the transpose Jacobian follows the reference implementation of \citet{meulemans2021dfc}; the feedback pathway is therefore computed from the forward network. The error that enters the controller is a scaled squared-error gradient at the output, so DFC propagates a filtered loss gradient backward through the forward network, and its update differs from the loss gradient. DFC shows the same delayed generalization as the reference (Figure~\ref{fig:grok}). Evaluated every $10$ epochs, training accuracy reaches $1.0$ at epochs $90$ to $130$ for DFC and at epoch $100$ for the reference, and test accuracy first exceeds $0.95$ at epochs $650$ to $710$ for DFC and $1220$ to $1450$ for the reference; the grokking delays are $520$, $600$ and $610$ epochs for DFC and $1120$ to $1350$ for the reference over the three seeds, all above the $500$-epoch criterion of Section~\ref{sec:setup}.

\paragraph{Measuring credit participation.}
\label{app:credit_measure}
Credit participation applies Equation~\ref{eq:pr} to the row norms of a neuron-level learning signal for the first hidden layer, one row per neuron, computed at every epoch of the last $800$ epochs and then averaged; it is the mean of instantaneous participation ratios. Which signal is logged depends on the condition (Table~\ref{tab:creditconv}). For the rules that hand a gradient-like signal to Adam it is that signal before clipping; for the adaptive optimizers, whose rescaling cannot be separated from their running state, it is the update applied to the first-layer weights. Circuit participation is measured from the outgoing weights of the trained network, on different parameters, so the agreement between credit and circuit participation is an empirical result and not an identity. Table~\ref{tab:baselines} also reports participation measured from Adam's effective step $m/\sqrt{v}$.

\begin{table}[!ht]
\caption{Signal on which credit participation is measured, per condition. All signals are for the first hidden layer, with one row per neuron.}
\label{tab:creditconv}
\begin{center}
\small
\begin{tabular}{@{}p{0.30\linewidth}p{0.66\linewidth}@{}}
\toprule
Condition & Logged signal \\
\midrule
Reference; LARS & raw loss gradient, before clipping and without the weight-decay term; for LARS this differs from the applied update only by the common layer factor and the weight-decay term \\
Credit equalization; group-size sweeps; imposed profiles & rescaled gradient of Equation~\ref{eq:knob} or~\ref{eq:grain}, before clipping and before Adam \\
DFC & local controller update of Equation~\ref{eq:update} for the first layer, before Adam \\
RMSProp, LAMB, Adafactor, Lion, Muon, Shampoo & update applied to the first-layer weights: gradient plus weight-decay term, transformed with the optimizer's running moments (Lion: sign of the momentum-interpolated gradient; LAMB: including the trust ratio) \\
\bottomrule
\end{tabular}
\end{center}
\end{table}

\section{Circuit measurements and algorithm checks}
\label{app:measures}

\subsection{Do outgoing norms reflect the distribution of computation?}
\label{app:functional}
We use the participation ratio of outgoing norms rather than a Fourier-based sparsity measure because it needs no Fourier basis, applies unchanged to every task we study, and is invariant to a uniform rescaling of the outgoing weights. It is, however, a structural measure. The network's function is unchanged when a ReLU neuron's incoming weights and bias are multiplied by a positive constant and its outgoing weights are divided by the same constant, so outgoing norms could in principle change without any change in the computation. We check the redistribution with two measures that are tied to the computation.

The first weights each neuron's outgoing norm by its mean absolute activation, $\mathbb{E}_x|h_j(x)| \, \|\mW^{\mathrm{out}}_{:,j}\|_2$, where $h_j(x)$ is the activation of neuron $j$ on input $x$ and the mean runs over the test inputs; this product is invariant to the compensating rescaling above. The second removes each neuron $j$ in turn and records $|\mathbb{E}_x[z_y(x) - z_y^{(-j)}(x)]|$, the absolute value of the mean change of the correct-class logit $z_y$ over the same inputs, where $z_y^{(-j)}$ is the logit with neuron $j$ removed; we apply Equation~\ref{eq:pr} to these ablation effects. Over eight seeds, activation-weighted participation increases from $0.343$ under the reference to $0.813$ under full credit equalization, and ablation participation from $0.422$ to $0.801$. The mean absolute change of the logit, $\mathbb{E}_x|z_y(x) - z_y^{(-j)}(x)|$, and of the margin give the same separation: $0.34$ to $0.49$ under the reference against $0.76$ to $0.98$ under equalization for the logit, and $0.38$ to $0.48$ against $0.85$ to $0.97$ for the margin (three seeds).

Low-norm neurons account for only part of the difference. We recompute participation among the neurons whose outgoing norm is at least one tenth of the layer mean. The reference has $84$ neurons below this threshold and participation $0.889\,(0.006)$ among the rest; full equalization has no neuron below the threshold and participation $0.991\,(0.001)$. Under equalization no neuron falls below the threshold, and the norms above it are also more uniform. It does not make the neurons less selective: the mean fraction of a neuron's input-weight Fourier power at its strongest frequency rises from $0.667$ under the reference to $0.884$ at $\beta = 1$ (five seeds, the experiment of Table~\ref{tab:suff}).

\subsection{Training fit and output scale}
\label{app:norm}
Participation could change because the training fit or the output scale changes. We therefore report the parameter norm $\|\theta\|$ and the logit margin $\gamma = \mathbb{E}_x[z_y(x) - \max_{c \neq y} z_c(x)]$, where $z_c(x)$ is the logit of class $c$ for input $x$, $y$ is the correct label of $x$ and the mean runs over the training set, for the reference, full credit equalization and DFC (Table~\ref{tab:norms}). The reference and the equalized network have nearly the same norm and margin despite their different participation, so the redistribution is not accompanied by a change in fit or output scale. DFC has a much smaller norm and margin: its outputs live on a different scale, which is one reason we use the credit intervention rather than DFC for the matched comparison.

\begin{table}[!ht]
\caption{Parameter norm and logit margin at matched weight decay (modular addition, $p=31$, four seeds; means and standard deviations). All runs have test accuracy $\geq 0.97$. Credit equalization leaves both quantities close to the reference values while producing a circuit at least as distributed as DFC's.}
\label{tab:norms}
\begin{center}
\begin{tabular}{lcc}
\toprule
Rule & $\|\theta\|$ & margin $\gamma$ \\
\midrule
Reference (backpropagation with Adam) & $26.10 \pm 0.15$ & $5.02 \pm 0.02$ \\
Credit equalization ($\beta{=}1$) & $26.65 \pm 0.10$ & $5.15 \pm 0.03$ \\
DFC                & $13.09 \pm 0.02$ & $0.54 \pm 0.00$ \\
\bottomrule
\end{tabular}
\end{center}
\end{table}

\subsection{Checks of the modular-addition computation}
\label{app:algo}
Prior work derives the computation that grokked networks use for modular addition: the first layer represents each operand by a few frequencies $k$, hidden neurons combine them into $\cos$ and $\sin$ of $k(a+b)$, and the logit for candidate $c$ sums $\cos(2\pi k(a+b-c)/p)$ over frequencies, which peaks at $c = a + b$ \citep{nanda2023progress,gromov2023grokking,chughtai2023toy,morwani2024feature}. We check whether the reference, DFC and full credit equalization all implement this computation, and then repeat the checks for the remaining grokking rules.

\paragraph{Fourier measurements.} Let $\vu_j, \vv_j \in \R^p$ be the halves of neuron $j$'s incoming weights that act on the two operands, and let $\hat{u}_j, \hat{v}_j$ be their discrete Fourier transforms. The power of neuron $j$ at frequency $k = 1, \dots, \lfloor p/2 \rfloor$ is
\begin{equation}
s_j(k) \;=\; |\hat{u}_j(k)|^2 + |\hat{v}_j(k)|^2,
\qquad
\hat{u}_j(k) = \textstyle\sum_{a=0}^{p-1} u_j[a]\, e^{-2\pi i k a / p}.
\label{eq:spectrum}
\end{equation}
The aggregate spectrum of a layer sums $s_j(k)$ over neurons.

\paragraph{Frequency ablation.} We keep only the strongest input frequencies, ranked by aggregate power, and count how many are needed to recover $95\%$ of the full test accuracy: $6.7$ for backpropagation with Adam, our example of a concentrated circuit, and $9.3$ for DFC, our example of a distributed circuit. Both networks rely on a small set of input frequencies. A comparison between the reference and full credit equalization, with a random control, follows.

\begin{figure}[!ht]
\begin{center}\includegraphics[width=0.8\linewidth]{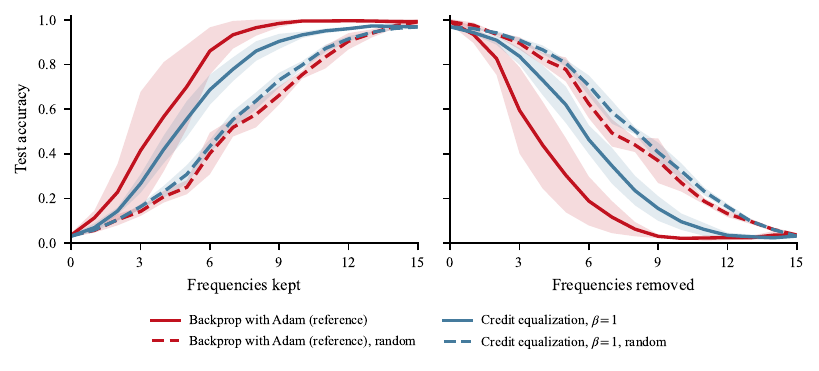}\end{center}
\caption{Test accuracy after keeping (left) or removing (right) $k$ input frequencies, ranked by first-layer weight power (solid) or drawn at random (dashed). Means over five seeds, bands one standard deviation across seeds.}
\label{fig:freqcurve}
\end{figure}

\paragraph{Frequency retention and removal under credit equalization.} We repeat the frequency ablation for the reference and full credit equalization over five seeds and add removal and a random control. Frequencies are ranked once per network by the summed power of its first-layer weights, so the ranking uses no data. We keep or remove the top $k$ frequencies for every $k$ from $0$ to $15$, always keep the constant component, and filter both operand blocks alike. The control keeps or removes $k$ frequencies drawn at random, averaged over $20$ draws per network. For both rules, keeping the top frequencies preserves more test accuracy than keeping random ones, and removing them costs more than removing random ones, at every $k$ from $2$ to $14$ in every seed (Table~\ref{tab:freqcurve}, Figure~\ref{fig:freqcurve}). The equalized network depends on its leading frequencies as the reference does, and spreads that dependence over more of them. Fitted on the training pairs only, the trigonometric function alone classifies every held-out pair correctly in every seed for both rules, with a mean margin between $6.1$ and $6.8$.

\begin{table}[!ht]
\caption{Test accuracy after keeping or removing $k$ input frequencies, ranked by first-layer weight power (top) or drawn at random ($20$ draws per network), modular addition at $p = 31$. Means over five seeds; the complete curves are in Figure~\ref{fig:freqcurve}.}
\label{tab:freqcurve}
\begin{center}
\small
\begin{tabular}{llcccccc}
\toprule
 & & \multicolumn{6}{c}{Frequencies kept or removed, $k$} \\
 & & $2$ & $4$ & $6$ & $8$ & $10$ & $12$ \\
\midrule
Reference & Keep top $k$ & $0.23$ & $0.57$ & $0.86$ & $0.97$ & $1.00$ & $1.00$ \\
 & Keep random $k$ & $0.10$ & $0.21$ & $0.40$ & $0.58$ & $0.76$ & $0.90$ \\
 & Remove top $k$ & $0.83$ & $0.44$ & $0.19$ & $0.06$ & $0.02$ & $0.02$ \\
 & Remove random $k$ & $0.94$ & $0.83$ & $0.62$ & $0.44$ & $0.27$ & $0.13$ \\
\midrule
Equalized & Keep top $k$ & $0.14$ & $0.42$ & $0.69$ & $0.86$ & $0.93$ & $0.96$ \\
 & Keep random $k$ & $0.10$ & $0.23$ & $0.43$ & $0.64$ & $0.80$ & $0.92$ \\
 & Remove top $k$ & $0.91$ & $0.73$ & $0.46$ & $0.23$ & $0.10$ & $0.03$ \\
 & Remove random $k$ & $0.94$ & $0.87$ & $0.71$ & $0.50$ & $0.32$ & $0.16$ \\
\bottomrule
\end{tabular}
\end{center}
\end{table}

\paragraph{Trigonometric fit.} We regress the full logit tensor onto the $(a+b-c)$ trigonometric basis of the Fourier algorithm. The fit explains $0.57$ of the logit variance for the reference, $0.42$ for full credit equalization and $0.90$ for DFC. The residual of the reference agrees across initializations with a correlation of only $0.235$, so there is little evidence of a second computation shared across seeds.

\paragraph{Agreement across runs.} With a fixed data split, seeds of the same rule differ only in their initialization, and their agreement is the yardstick for comparisons across rules. Logit tensors of reference seeds correlate at $0.446$; correlations across rules lie between $0.427$ and $0.467$. Linear centered kernel alignment between the first-hidden-layer activations of the reference and of full equalization is $0.69$, compared with $0.57$ between reference seeds and $0.74$ between equalized seeds. Agreement across rules is thus comparable to agreement across seeds of one rule.

\paragraph{The published alternative.} The pizza algorithm of \citet{zhong2023clock} multiplies the clock logit by an envelope that depends on the difference of the operands, $\lvert\cos(\pi k(a-b)/p)\rvert \cdot \cos(2\pi k(a+b-c)/p)$. Its basis overlaps with the trigonometric one: on a synthetic pizza tensor the trigonometric basis explains $0.81$ of the variance, and on a synthetic clock tensor the pizza basis explains $0.81$, so explained variance alone does not tell the two apart. Two statistics do. Adding the pizza basis to the trigonometric basis raises the explained variance by at most $0.0012$ in every seed of every rule, against $0.19$ for a synthetic pizza tensor and $0.001$ for noise. The distance irrelevance of \citet{zhong2023clock}, the variability of the correct logit at a fixed operand difference relative to its total variability, lies between $0.976$ and $0.990$ for every seed of every rule; they report $0.85$ for a clock network and $0.17$ for a pizza network, and classify networks below $0.6$ as pizza. None of the rules implements this alternative.

\paragraph{The remaining grokking rules.} We repeat the checks for the other learning rules that grok: the eight optimizers of Table~\ref{tab:zoo}, trained as in Appendix~\ref{app:repro}, and the two constructed rules with a fresh or a fading orthogonal component, trained as in Appendix~\ref{app:generalization}, with three seeds each. As above, the seeds share one data split and differ in their initialization, and a reference trained in the same way provides the yardstick (Table~\ref{tab:algoall}). In every rule the trigonometric basis explains between $0.42$ and $0.68$ of the logit variance per seed, the distance irrelevance lies above $0.97$, and the logit correlation and the linear centered kernel alignment of first-hidden-layer activations with the reference across initializations lie at the level of the reference's agreement with itself. Together, the checks show the same periodic computation across all tested grokking rules and distinguish it from the published alternative.

\begin{table}[!ht]
\caption{Algorithm checks for the remaining grokking rules (modular addition, $p=31$, three seeds, one shared data split). Trig: fraction of logit variance explained by the $(a+b-c)$ basis. Distance irrelevance: the statistic of \citet{zhong2023clock}, above $0.6$ for clock networks and below $0.6$ for pizza networks. Correlation and CKA: agreement with the reference trained in the same way, across initializations; the reference rows give its agreement with itself. The optimizers are trained for $20{,}000$ epochs and the constructed rules for $15{,}000$, each with its own reference. The seeds share one data split, so the values differ slightly from Table~\ref{tab:zoo}, whose seeds use different splits.}
\label{tab:algoall}
\begin{center}
\small
\setlength{\tabcolsep}{3pt}
\begin{tabular}{lcccccc}
\toprule
Rule & Test & Circuit PR & Trig & Dist. irrel. & Correlation & CKA \\
\midrule
\multicolumn{7}{l}{\emph{Optimizers, $20{,}000$ epochs}} \\
Reference & $0.999$ & $0.558$ & $0.535$ & $0.983$ & $0.407$ & $0.590$ \\
LARS & $0.975$ & $0.604$ & $0.550$ & $0.986$ & $0.419$ & $0.592$ \\
RMSProp & $0.995$ & $0.496$ & $0.517$ & $0.984$ & $0.456$ & $0.598$ \\
LAMB & $0.998$ & $0.566$ & $0.463$ & $0.984$ & $0.396$ & $0.573$ \\
Adafactor & $0.998$ & $0.489$ & $0.553$ & $0.983$ & $0.432$ & $0.580$ \\
Lion & $0.998$ & $0.479$ & $0.558$ & $0.983$ & $0.461$ & $0.585$ \\
Muon & $0.993$ & $0.494$ & $0.498$ & $0.980$ & $0.407$ & $0.554$ \\
Shampoo & $0.992$ & $0.529$ & $0.525$ & $0.980$ & $0.339$ & $0.518$ \\
Nero & $0.988$ & $0.986$ & $0.455$ & $0.984$ & $0.435$ & $0.659$ \\
\midrule
\multicolumn{7}{l}{\emph{Constructed rules, $15{,}000$ epochs}} \\
Reference & $0.998$ & $0.579$ & $0.532$ & $0.983$ & $0.399$ & $0.594$ \\
Fresh noise & $0.993$ & $0.571$ & $0.531$ & $0.984$ & $0.467$ & $0.621$ \\
Error-gated bias & $0.984$ & $0.688$ & $0.458$ & $0.980$ & $0.415$ & $0.603$ \\
\bottomrule
\end{tabular}
\end{center}
\end{table}

\section{Controls on credit equalization}
\label{app:credit_controls}
The main experiment changes the credit magnitudes and observes a change in the outgoing weights. This section tests the alternative explanations: that the effect comes from the incoming weight norms rather than from the credit, that it depends on which neurons receive the credit, that Adam or the hyperparameters produce it, and that the concentrated circuit is only a transient state.

\subsection{Constraining incoming weights}
\label{app:named}
Full credit equalization gives every neuron the same gradient norm, and Nero, which also distributes the circuit, additionally constrains the weight norms. Could equal incoming weight norms alone explain the redistribution? We test two constraints on the weights that leave the gradients unchanged. The first renormalizes each neuron's incoming weights to a fixed norm after every update. Circuit participation is $0.661\,(0.015)$, compared with $0.595\,(0.023)$ for the reference, and deletion tolerance decreases (Table~\ref{tab:named}, upper block). The second is weight normalization \citep{salimans2016weightnorm}, which parameterizes each incoming weight vector as a direction times a learned scale; with weight decay on the scales it groks to test accuracy $0.993$ with participation $0.466\,(0.026)$. Neither constraint approaches the participation of $0.99$ reached by credit equalization. Equal incoming weight norms do not reproduce the redistribution.

\subsection{Assigning credit independently of neuron identity}
\label{app:imposed}
In the first hidden layer, a neuron's incoming-weight gradient is the input multiplied by the backpropagated error, and that error depends linearly on the neuron's outgoing weights. Credit and outgoing norms are therefore coupled at every step, and their agreement could reflect this coupling rather than an effect of the credit on the circuit. We break the coupling by assigning credit magnitudes that carry no information about the neurons.

A \emph{credit profile} is a vector of $n$ nonnegative target magnitudes with a chosen participation ratio. At every step we replace the incoming-gradient norms by this profile, scaled so that the total norm is preserved, and keep every gradient's direction. A random permutation, drawn once per run, decides which neuron receives which magnitude, so the assignment is independent of the neurons' roles. Circuit participation follows the imposed value, from $0.637$ at profile participation $0.60$ to $0.990$ at $1.00$, deletion tolerance increases with it, and test accuracy stays above $0.96$ (Table~\ref{tab:named}, lower block). An allocation chosen independently of the neurons' current roles therefore sets the final concentration.

A second control keeps the reference's credit magnitudes at each step but randomly reassigns them to the neurons at every step. Each permutation preserves that step's participation and changes only which neurons receive the largest signals. Final circuit participation is $0.992$, compared with $0.613$ for the reference. Measured over the last $800$ epochs, credit participation is $0.979$ rather than the reference's $0.604$: once the assignment no longer favors the same neurons, the trained network itself develops more uniform gradients. This suggests that concentration requires the credit to stay concentrated on the same neurons over time.

The effect persists across the tested initialization scales. Across scales from $0.5$ to $4$, reference participation is $0.586$ to $0.613$ and full equalization gives $0.990$ to $0.998$, with a mean absolute difference of $0.008$ between credit and circuit participation. The effect is also local to the layer in which credit is equalized: equalizing only the second hidden layer gives participation $1.000$ there while the first layer stays at $0.593\,(0.004)$ with $86$ low-norm neurons, and equalizing both layers distributes both, at test accuracy of at least $0.98$.

\begin{table}[!ht]
\caption{Controls on the weights and on the credit profile ($p=31$, three seeds; standard deviations in parentheses). Upper block: constraining the incoming weight norms without rescaling the gradients leaves the circuit concentrated. Lower block: credit profiles of a chosen participation, assigned to neurons by a random permutation, produce circuit participation within $0.018$ of the imposed value on average; the final row reassigns the reference's credit magnitudes to random neurons at every step. Each block has its own reference run. Random del and norm-ranked del are deletion tolerances (Appendix~\ref{app:deletion}).}
\label{tab:named}
\begin{center}
\small
\setlength{\tabcolsep}{4pt}
\begin{tabular}{lccccc}
\toprule
Condition & Test & Credit PR & Circuit PR & Random del & Norm-ranked del \\
\midrule
\multicolumn{6}{l}{\emph{Constraining the incoming weights}} \\
Reference & $0.998$ & $0.586$ & $0.595\,(0.023)$ & $0.672$ & $0.311$ \\
Fixed incoming norm & $0.954$ & $0.611$ & $0.661\,(0.015)$ & $0.607$ & $0.256$ \\
\midrule
\multicolumn{6}{l}{\emph{Imposing a credit profile}} \\
Reference & $0.997$ & $0.604$ & $0.613\,(0.016)$ & $0.673$ & $0.298$ \\
Imposed $0.60$ & $0.980$ & $0.600$ & $0.637\,(0.003)$ & $0.607$ & $0.240$ \\
Imposed $0.70$ & $0.981$ & $0.700$ & $0.727\,(0.001)$ & $0.663$ & $0.319$ \\
Imposed $0.80$ & $0.972$ & $0.800$ & $0.815\,(0.005)$ & $0.693$ & $0.420$ \\
Imposed $0.90$ & $0.969$ & $0.900$ & $0.902\,(0.005)$ & $0.723$ & $0.533$ \\
Imposed $1.00$ & $0.975$ & $1.000$ & $0.990\,(0.003)$ & $0.775$ & $0.599$ \\
Reshuffled each step & $0.974$ & $0.979$ & $0.992\,(0.002)$ & $0.775$ & $0.601$ \\
\bottomrule
\end{tabular}
\end{center}
\end{table}

\subsection{Optimizer and hyperparameter controls}
\label{app:baselines}
Adam normalizes each parameter separately, so one might suspect that Adam either creates the redistribution or removes it. Neither is the case. With SGD with momentum in place of Adam, circuit participation is $0.53$ at $\beta = 0$ and $0.99$ at $\beta = 1$, with deletion tolerance $0.60$ and $0.76$ (Table~\ref{tab:baselines}). With Lion in place of Adam, full credit equalization gives participation $0.984\,(0.003)$ at test accuracy $0.974\,(0.010)$, whereas Lion alone gives about $0.5$. In the Adam runs, participation measured from Adam's effective step $m/\sqrt{v}$ is $0.66$ at $\beta = 0$ and $0.99$ at $\beta = 1$, so the difference between neurons survives Adam's transformation.

Coupled weight decay is added to the rescaled gradient, so equalization also changes the size of the decay term relative to each neuron's gradient, and Appendix~\ref{app:decay} shows that under Adam this accounts for the redistribution. Our Lion harness also adds the decay before the update. For SGD with coupled decay, rescaling a neuron's gradient by $s_j$ equals a per-neuron learning rate $s_j$ with decay $\lambda/s_j$, whose fixed points are those of the decay control, so both optimizers follow the same per-neuron condition. To test the intervention without this pathway, we repeat the comparison with AdamW, which applies weight decay to the weights directly. Weight decay $1$ is the only value in $\{10^{-3}, 10^{-2}, 10^{-1}, 1\}$ at which the AdamW reference groks within $20{,}000$ epochs. At that value, over $80{,}000$ epochs and five seeds measured every $5000$ epochs, equalization gives higher outgoing participation at all $80$ paired seed checkpoints, with $0.96$ against $0.91$ at the end of training. Activation-weighted and ablation participation are higher under equalization at $69$ and $74$ of the $80$ checkpoints. Both functional measures fluctuate strongly in both conditions. Over the last $30{,}000$ epochs they range from $0.02$ to $0.77$ for the reference and from $0.13$ to $0.98$ under equalization, and at the end of training they average $0.41$ and $0.36$ for the reference and $0.82$ and $0.73$ under equalization. Test accuracy lies between $0.94$ and $1.00$ in both conditions. The low functional values of the reference come from $20$ to $50$ first-layer neurons that become active on every input and add a constant offset to all logits, which zero-ablation counts as large effects. Replacing these neurons by their mean activation changes test accuracy by at most $0.003$. Computed over the other neurons and without the constant part of each activation, activation-weighted participation is $0.88$ for the reference and $0.97$ under equalization, higher in all ten seeds. The AdamW reference is thus distributed in both kinds of measure, and equalization distributes it further.

Ordinary tuning of the reference does not distribute the circuit either. Across three learning rates and two weight-decay values, every setting groks with circuit participation between $0.530$ and $0.626$ (Table~\ref{tab:bpgrid}). Credit participation ranges from $0.523$ to $0.613$ and differs from circuit participation by $0.009$ on average.

Dropout is the natural comparison for deletion tolerance, since it trains the network under random deletion. It distributes the circuit as well, to participation $0.95$ at rate $0.1$ and $0.99$ at rate $0.5$, and its random-deletion tolerance, $0.79$ and $0.82$, is close to the $0.77$ of full credit equalization, which never sees a deleted neuron during training. Two unrelated interventions thus reach the same participation and nearly the same tolerance, which supports the reading of Appendix~\ref{app:suff} that, in this network, deletion tolerance follows the distribution of outgoing norms however that distribution is produced. The interventions differ in what else they change. Dropout modifies the objective and raises the margin and the parameter norm (Table~\ref{tab:baselines}), whereas credit equalization changes only the learning rule and leaves loss, margin and norm within a few percent of the reference. Intermediate equalization strengths lie between the endpoints: at $\beta = 0.5$, circuit participation is $0.98$ and test accuracy $0.985$; at $\beta = 0.75$, they are $0.99$ and $0.977$.

\begin{table}[!ht]
\caption{Optimizer and regularizer controls ($p=31$, three seeds). Upper block: circuit participation and random-deletion tolerance under other optimizers and under dropout; margin and norm are given where measured. Lower block: participation ratio of the measured learning signal in the Adam runs, from the raw gradient and from Adam's effective step $m/\sqrt{v}$.}
\label{tab:baselines}
\begin{center}
\small
\begin{tabular}{lcccc}
\toprule
Setting & Circuit PR & Random del & Margin & Norm \\
\midrule
\multicolumn{5}{l}{\emph{Circuit participation}} \\
SGD with momentum, $\beta{=}0$ & $0.53$ & $0.60$ & --- & --- \\
SGD with momentum, $\beta{=}1$ & $0.99$ & $0.76$ & --- & --- \\
Lion, $\beta{=}1$ & $0.984\,(0.003)$ & --- & --- & --- \\
Dropout $0.1$ & $0.95$ & $0.79$ & $5.3$ & $26.8$ \\
Dropout $0.5$ & $0.99$ & $0.82$ & $6.8$ & $30.2$ \\
Credit equalization, $\beta{=}1$ (Adam) & $0.99$ & $0.77$ & $5.2$ & $26.6$ \\
\midrule
\multicolumn{5}{l}{\emph{Participation of the measured learning signal (Adam runs)}} \\
Raw gradient, $\beta{=}0$ & $0.59$ & --- & --- & --- \\
Effective step $m/\sqrt{v}$, $\beta{=}0$ & $0.66$ & --- & --- & --- \\
Effective step $m/\sqrt{v}$, $\beta{=}1$ & $0.99$ & --- & --- & --- \\
\bottomrule
\end{tabular}
\end{center}
\end{table}

\begin{table}[!ht]
\caption{Learning-rate and weight-decay grid for the reference ($p=31$, three seeds; standard deviations in parentheses). All settings grok and retain concentrated circuits, with credit participation close to circuit participation.}
\label{tab:bpgrid}
\begin{center}
\small
\begin{tabular}{lcccc}
\toprule
Learning rate & Weight decay & Test & Credit PR & Circuit PR \\
\midrule
$3\times10^{-4}$ & $10^{-3}$        & $0.998$ & $0.613$ & $0.626\,(0.012)$ \\
$3\times10^{-4}$ & $3\times10^{-3}$ & $0.983$ & $0.560$ & $0.571\,(0.027)$ \\
$10^{-3}$        & $10^{-3}$        & $0.997$ & $0.604$ & $0.613\,(0.016)$ \\
$10^{-3}$        & $3\times10^{-3}$ & $0.993$ & $0.523$ & $0.530\,(0.008)$ \\
$3\times10^{-3}$ & $10^{-3}$        & $0.993$ & $0.589$ & $0.594\,(0.024)$ \\
$3\times10^{-3}$ & $3\times10^{-3}$ & $0.988$ & $0.546$ & $0.553\,(0.022)$ \\
\bottomrule
\end{tabular}
\end{center}
\end{table}

\subsection{Longer training and switching the learning rule}
\label{app:long}
Is the concentrated reference circuit a stable solution, or a stage on the way to a distributed one? We extend training from $20{,}000$ to $200{,}000$ epochs. Reference participation stabilizes by approximately $40{,}000$ epochs and ends at $0.605\,(0.012)$; full equalization ends at $0.990\,(0.003)$ (Table~\ref{tab:long}). Longer training does not distribute the reference circuit.

We then train for $20{,}000$ epochs with one rule and continue for $30{,}000$ epochs with the other, with same-rule continuations as controls. Switching from the reference to full equalization raises participation from $0.613\,(0.016)$ to $0.991\,(0.000)$ within $5000$ epochs, at test accuracy $0.997$; continuing with the reference leaves it at $0.610\,(0.012)$. The reverse switch lowers participation from $0.990\,(0.003)$ to $0.722\,(0.016)$ and is still decreasing when the run ends, while its control stays at $0.990\,(0.003)$. Task and data are unchanged in all four runs; only the credit assignment changes. A fixed rule thus preserves its circuit over the tested horizon, and changing the credit allocation reorganizes the outgoing weights of an already learned circuit in either direction. Figure~\ref{fig:steer}b shows a separate run of the same protocol at $100$-epoch resolution: after the switch to equalization, participation exceeds $0.95$ within $300$ epochs, and after the reverse switch it reaches $0.67$ after $30{,}000$ epochs and is still decreasing.

\begin{table}[!ht]
\caption{Circuit stability under a fixed rule and reorganization after switching ($p=31$, three seeds; standard deviations in parentheses). Fixed-rule runs show participation after $20{,}000$ and $200{,}000$ epochs. Switched runs use $20{,}000$ epochs in the first phase and $30{,}000$ in the second and show participation after each phase; the reverse switch has not stabilized by the end. Same-rule continuations control for the additional training.}
\label{tab:long}
\begin{center}
\small
\begin{tabular}{lccc}
\toprule
Condition & Circuit PR before & Circuit PR after & Test \\
\midrule
\multicolumn{4}{l}{\emph{Rule held, $20{,}000$ to $200{,}000$ epochs}} \\
Reference ($\beta{=}0$) & $0.60$ & $0.605\,(0.012)$ & $0.997$ \\
Credit equalization ($\beta{=}1$) & $0.99$ & $0.990\,(0.003)$ & $0.975$ \\
\midrule
\multicolumn{4}{l}{\emph{Rule switched after phase one}} \\
$\beta{=}0 \to \beta{=}1$ & $0.613\,(0.016)$ & $0.991\,(0.000)$ & $0.997$ \\
$\beta{=}1 \to \beta{=}0$ & $0.990\,(0.003)$ & $0.722\,(0.016)$ & $0.990$ \\
$\beta{=}0 \to \beta{=}0$ (control) & $0.613\,(0.016)$ & $0.610\,(0.012)$ & $0.997$ \\
$\beta{=}1 \to \beta{=}1$ (control) & $0.990\,(0.003)$ & $0.990\,(0.003)$ & $0.975$ \\
\bottomrule
\end{tabular}
\end{center}
\end{table}

\subsection{When the circuit concentrates}
\label{app:timing}
We train the reference (modular addition, $p = 31$, five fresh seeds) for $20{,}000$ epochs and record circuit participation every $50$ epochs. All runs fit the training set by epoch $100$, when participation is $0.997$ or higher. Test accuracy first reaches $0.95$ between epochs $1250$ and $1500$, and participation has fallen to between $0.59$ and $0.67$ by then. From the grokking epoch to the end of training it changes by $-0.04$ to $+0.05$, while the training loss settles at about $0.04$. The concentration of the circuit thus forms during the grokking delay, together with the generalizing computation, and not in a later phase of training.

\subsection{Gradient and weight decay under Adam}
\label{app:decay}
Our reference adds the coupled decay term $\lambda \vw_j$ to each neuron's incoming-weight gradient $\vg_j$ before Adam, whose update is nearly invariant to a constant rescaling of its input. Full credit equalization multiplies $\vg_j$ by $s_j = \bar{g}/\|\vg_j\|_2$ and leaves the decay term unchanged, so it also changes the ratio between the two. We separate the two effects with four conditions that differ only in how each first-layer neuron's gradient and decay term are scaled before Adam, with $s_j$ recomputed at every step (Table~\ref{tab:decay}). The decay is added after clipping, where Adam adds coupled decay, and biases and the other layers keep the decay $\lambda$. Scaling only the decay term, by $1/s_j$, gives the participation of full equalization, $0.993$, no nearly unused neurons and nearly the same deletion tolerance, and it is more distributed still on the activation-weighted and ablation measures ($0.981$ and $0.924$ against $0.850$ and $0.832$). Scaling both the gradient and the decay term by $s_j$ keeps each neuron's ratio of gradient to decay as in the reference and gives the reference circuit, with participation $0.574$ and $92$ nearly unused neurons. Under Adam with coupled weight decay, the intervention therefore acts through this ratio. Adam's invariance holds exactly only for a constant factor, and $s_j$ changes from step to step, so the two conditions do not give identical updates; the controls measure their effect on the circuit directly.

\paragraph{Where the three non-gradient terms enter the update.} For the incoming row $\vw_j$ of neuron $j$ with loss gradient $\vg_j$ and learning rate $\eta$, Adam with coupled decay forms its moment estimates $\hat{\vm}_j$ and $\hat{\vv}_j$ from $\vg_j + \lambda \vw_j$ and steps
\begin{equation}
\Delta \vw_j = -\eta\, \hat{\vm}_j \big/ \big(\sqrt{\hat{\vv}_j} + \epsilon\big).
\end{equation}
Adam removes a constant overall scale of its input, so a neuron whose gradient is small relative to $\lambda \vw_j$ still receives a step of full size, and that step mostly shrinks its weights. AdamW forms the moments from $\vg_j$ alone and applies the decay outside,
\begin{equation}
\Delta \vw_j = -\eta\, \big[\hat{\vm}_j \big/ \big(\sqrt{\hat{\vv}_j} + \epsilon\big) + \lambda \vw_j\big].
\end{equation}
While $\sqrt{\hat{\vv}_j} \gg \epsilon$ the first term does not depend on the scale of the gradient. Once $\sqrt{\hat{\vv}_j} \ll \epsilon$, as in the wide transformer of Appendix~\ref{app:transformer}, it is approximately $\hat{\vm}_j / \epsilon$, so the gradient competes with $\epsilon \lambda \vw_j$, and $\epsilon$ sets the scale against which the gradient is compared, as $\lambda$ does under coupled decay. A trainable gain $\gamma_j$ and the outgoing column $\va_j$ are updated by the first rule with their own gradients, so each has its own ratio of gradient to decay, which rescaling $\vg_j$ does not change (Appendix~\ref{app:normlayers}). In each case the controls of Table~\ref{tab:pathways} multiply the term that does not scale with the gradient, $\lambda$, $\epsilon$ or the decay of $\gamma_j$ and $\va_j$, by $1/s_j$, or multiply the gradient and this term together by $s_j$.

\begin{table}[!ht]
\caption{Gradient and coupled decay under Adam (modular addition, $p=31$, five seeds, $20{,}000$ epochs; standard deviations in parentheses). $s_j$ is the factor of full credit equalization for neuron $j$, recomputed at every step; the first two rows are the reference and full credit equalization. Nearly unused neurons have an outgoing norm below a tenth of the layer mean. Act.\ and abl.\ are the activation-weighted and ablation participation of Appendix~\ref{app:functional}; random and norm-ranked are deletion tolerances (Appendix~\ref{app:deletion}).}
\label{tab:decay}
\begin{center}
\footnotesize
\setlength{\tabcolsep}{3pt}
\begin{tabular}{llccccccc}
\toprule
Gradient & Decay & Test & Circuit PR & Nearly unused & Act. & Abl. & Random & Norm-ranked \\
\midrule
$\vg_j$ & $\lambda \vw_j$ & $0.995$ & $0.606\,(0.019)$ & $83.6$ & $0.355$ & $0.429$ & $0.675$ & $0.340$ \\
$s_j \vg_j$ & $\lambda \vw_j$ & $0.976$ & $0.993\,(0.001)$ & $0$ & $0.850$ & $0.832$ & $0.776$ & $0.585$ \\
$\vg_j$ & $\lambda \vw_j / s_j$ & $0.979$ & $0.993\,(0.001)$ & $0$ & $0.981$ & $0.924$ & $0.782$ & $0.611$ \\
$s_j \vg_j$ & $s_j \lambda \vw_j$ & $0.996$ & $0.574\,(0.046)$ & $92.4$ & $0.283$ & $0.378$ & $0.664$ & $0.316$ \\
\bottomrule
\end{tabular}
\end{center}
\end{table}

\section{Normalization groups and optimizer comparison}
\label{app:groups_opt}

\subsection{Normalizing groups of neurons}
\label{app:grain}
Section~\ref{sec:law} shows that the number of neurons sharing one normalization factor decides how far the circuit is redistributed. This section gives the details of that sweep. We divide the $256$ neurons of the first hidden layer into $g$ equal groups of $m = n/g$ neurons, fixed at the start of the run, and at every step multiply all gradients in a group by the common factor of Equation~\ref{eq:grain}, so that the mean credit is equal across groups while the relative magnitudes within a group are unchanged; a group whose gradients are all zero is left unchanged. Group size $m = 256$ corresponds to no rescaling and reproduces the reference: $0.602\,(0.023)$ against $0.60\,(0.02)$ in the $\beta$ sweep. Group size $m = 1$ rescales each neuron separately and reproduces full credit equalization: $0.992$ against $0.99$ at $\beta = 1$. Across the per-neuron settings, credit and circuit participation differ by $0.0074$ on average within seeds, which is the vertical distance from the identity line in Figure~\ref{fig:law}a.

Most of the change happens at the smallest groups. Going from pairs to single neurons raises circuit participation from $0.753$ to $0.992$, about $61\%$ of the total increase from the whole-layer value, and norm-ranked deletion tolerance from $0.327$ to $0.576$. Groups of four and two give intermediate values. Per-neuron rescaling of the credit is thus the setting most likely to distribute a circuit, and coarser groups keep most of the concentration.

On $S_5$ the sweep gives $0.749$ for groups of four, $0.852$ for pairs and $0.974$ for single neurons, the same pattern as on modular addition.

Adam-mini \citep{zhang2025adammini} uses one second-moment estimate per row of each weight matrix, that is, per output neuron. With coupled decay it normalizes each neuron's gradient and decay term together, so their ratio stays as in Adam. On modular addition with the reference settings and five fresh seeds, its circuit participation is $0.586$ against $0.594$ for Adam, and greedy pruning keeps $72$ neurons against $70$ at test accuracy $0.95$ (Appendix~\ref{app:minimal}). Per-neuron normalization distributes the circuit only when it changes each neuron's credit relative to its decay.

\subsection{Comparison across optimizers}
\label{app:zoo}
The controlled sweep varies only the group size and preserves the direction of every neuron's gradient before Adam. Optimizers combine a group structure with other transformations, so we ask whether they nevertheless sort by their normalization group. Table~\ref{tab:zoo} reports test accuracy, credit participation and circuit participation for the eight optimizers of Table~\ref{tab:ruleset} with the settings of Appendix~\ref{app:repro}. LARS (one factor per layer), RMSProp and Lion (per parameter), LAMB (per parameter with a layerwise trust ratio), Adafactor (factored per-parameter scaling), and Muon and Shampoo (matrix transformations) all produce concentrated circuits, with participation between $0.470$ and $0.613$, although their credit participation ranges from $0.478$ to $1.000$. Nero normalizes per neuron and produces a distributed circuit. At its default learning rate $10^{-2}$ and at $3\times10^{-3}$, it generalizes with test accuracy $0.985$ and $0.978$ and participation $0.986\,(0.002)$ and $0.983\,(0.001)$ (lower block of Table~\ref{tab:zoo}). It also tolerates neuron deletion better than the reference: random-deletion tolerance is $0.748$ and norm-ranked tolerance $0.626$ ($0.728$ and $0.541$ at the smaller learning rate), against $0.672$ and $0.311$ for the reference in the same experiment, close to the values of the equalized network. A per-neuron optimizer thus provides the same robustness as our intervention. Because Nero holds each neuron's incoming-weight norm fixed, its outgoing-norm participation cannot arise from scale moving between a neuron's incoming and outgoing weights, the rescaling symmetry discussed in Appendix~\ref{app:functional}.

Proposition~\ref{prop:grain} covers positive rescalings that preserve update directions, whereas per-parameter and matrix transformations rotate the update, so the optimizer comparison extends the controlled sweep empirically. NorMuon \citep{li2025normuon} adds a per-neuron normalization after Muon's orthogonalization, motivated by the unequal neuron update norms that Muon produces, the same inequality that concentrates the circuit here.

\begin{table}[!ht]
\caption{Optimizer comparison on modular addition ($p=31$, three seeds, $20{,}000$ epochs). The upper block runs every optimizer at the learning rate $10^{-3}$ and weight decay $10^{-3}$ of the reference (LARS at fifty times the learning rate); the reference and full credit equalization are included as comparison points. The lower block reports Nero, which uses no weight decay, at its default learning rate $10^{-2}$ and at $3\times10^{-3}$. Credit participation is measured on the raw gradient for the reference and LARS, on the rescaled gradient for credit equalization, and on the applied update for the other optimizers (Appendix~\ref{app:repro}). Optimizers: LARS \citep{you2017lars}, RMSProp \citep{tieleman2012rmsprop}, LAMB \citep{you2020lamb}, Adafactor \citep{shazeer2018adafactor}, Lion \citep{chen2023lion}, Muon \citep{jordan2024muon}, Shampoo \citep{gupta2018shampoo}, Nero \citep{liu2021nero}.}
\label{tab:zoo}
\begin{center}
\footnotesize
\setlength{\tabcolsep}{4pt}
\begin{tabular}{llccc}
\toprule
Optimizer & Normalizes & Test & Credit PR & Circuit PR \\
\midrule
Reference (backpropagation, Adam) & per parameter & $0.998$ & $0.586$ & $0.595$ \\
LARS            & per layer     & $0.967$ & $0.520$ & $0.613$ \\
RMSProp         & per parameter & $0.992$ & $0.791$ & $0.470$ \\
LAMB            & per parameter and per layer & $0.995$ & $0.779$ & $0.543$ \\
Adafactor       & factored, per parameter & $0.991$ & $0.839$ & $0.518$ \\
Lion            & per parameter (sign) & $0.985$ & $1.000$ & $0.535$ \\
Muon            & per matrix    & $0.999$ & $0.478$ & $0.524$ \\
Shampoo         & per matrix    & $0.986$ & $0.521$ & $0.555$ \\
Credit equalization ($\beta{=}1$) & per neuron & $0.976$ & $1.000$ & $0.992$ \\
\midrule
\multicolumn{5}{l}{\emph{Nero, no weight decay}} \\
Nero, learning rate $10^{-2}$ & per neuron & $0.985$ & --- & $0.986\,(0.002)$ \\
Nero, learning rate $3\times10^{-3}$ & per neuron & $0.978$ & --- & $0.983\,(0.001)$ \\
\bottomrule
\end{tabular}
\end{center}
\end{table}

\section{Neuron deletion and adaptation}
\label{app:consequences}

\subsection{Neuron deletion}
\label{app:deletion}
Deleting a first-hidden-layer neuron sets its outgoing weights to zero; the network is then evaluated on the test set without retraining. Random deletion removes fractions $0, 0.1, \ldots, 0.8$ of the layer, with $20$ random masks per fraction ($30$ in the ten-seed experiment of Figure~\ref{fig:main}c). Deletion tolerance is the area under the curve of test accuracy against the deleted fraction at these nine fractions, computed by the trapezoidal rule and divided by the range $0.8$, so that it lies between $0$ and $1$. Norm-ranked deletion removes neurons in decreasing order of outgoing norm at the same fractions. Greedy deletion evaluates every remaining neuron, removes the one whose removal lowers test accuracy the most, and repeats until accuracy falls below $0.5$; we report the number of neurons removed. Greedy deletion selects neurons by their effect on the test set rather than by their weights.

\begin{figure}[!ht]
\begin{center}\includegraphics[width=0.9\linewidth]{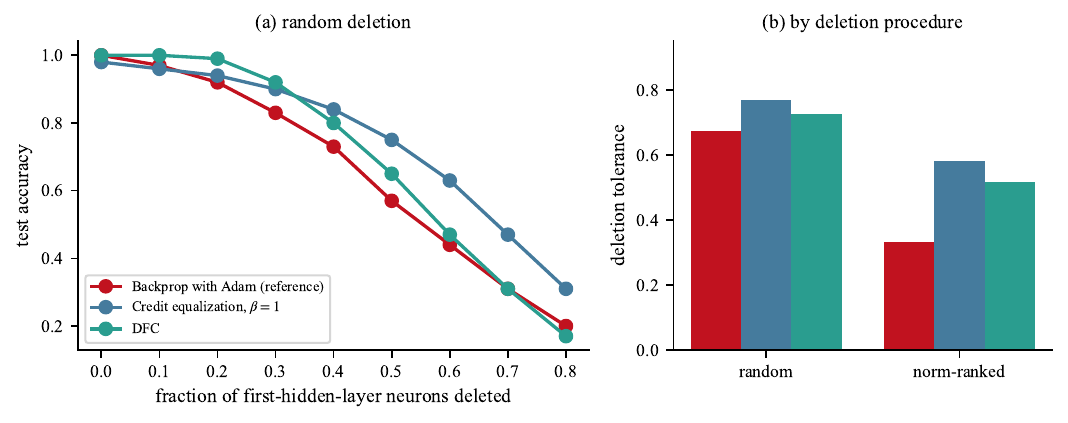}\end{center}
\caption{Neuron deletion for the reference, full credit equalization ($\beta=1$) and DFC on modular addition ($p=31$, three seeds). \textbf{(a)} Test accuracy against the fraction of randomly deleted first-hidden-layer neurons. \textbf{(b)} Deletion tolerance, the normalized area under the accuracy curve over the deletion fractions, under random and norm-ranked deletion.}
\label{fig:fragile}
\end{figure}

Table~\ref{tab:targeted} compares the reference, full credit equalization and DFC under random and norm-ranked deletion in one matched three-seed experiment (Figure~\ref{fig:fragile}). Equalization improves tolerance under both procedures, and the improvement is larger under norm-ranked deletion, which removes exactly the neurons a concentrated circuit relies on. DFC lies between the two. Table~\ref{tab:plane} reports greedy deletion for the reference and three equalization strengths: the number of neurons that must be removed grows with participation, from $23$ at $\beta = 0$ to $61$ at $\beta = 1$. The deletion advantage of the distributed circuit thus does not depend on selecting neurons by their norm. Over ten seeds, random-deletion tolerance is $0.672\,(0.012)$ for the reference and $0.773\,(0.014)$ under full equalization (Figure~\ref{fig:main}c).

\paragraph{Removal in retention order.} Pruning removes neurons in the order that keeps performance. We remove first-hidden-layer neurons in increasing order of outgoing norm (magnitude pruning) and, greedily, the neuron whose removal raises the training cross-entropy least, and record test accuracy without retraining (Table~\ref{tab:retain}; five seeds, the networks of Table~\ref{tab:decay}). The concentrated circuit keeps test accuracy above $0.9$ until $62\%$ of its neurons are removed by magnitude, against $20\%$ for the distributed circuit. Greedy removal takes away $191$ against $160$ neurons before test accuracy falls below $0.9$, and $177$ against $117$ before it falls below $0.95$. The nearly unused neurons of the concentrated circuit can be removed at no cost, so it is the easier circuit to compress.

\begin{table}[!ht]
\caption{Removal in retention order (modular addition, $p=31$, five seeds; standard deviations in parentheses). Magnitude pruning removes neurons in increasing order of outgoing norm, evaluated at fractions $0, 0.1, \ldots, 0.8$; greedy pruning removes the neuron whose removal raises the training cross-entropy least.}
\label{tab:retain}
\begin{center}
\small
\begin{tabular}{lcc}
\toprule
 & Concentrated ($\beta=0$) & Distributed ($\beta=1$) \\
\midrule
Magnitude pruning, tolerance & $0.949$ & $0.718$ \\
Magnitude pruning, largest fraction with test $\geq 0.9$ & $0.62\,(0.04)$ & $0.20\,(0.00)$ \\
Greedy pruning, neurons removed with test $\geq 0.9$ & $191\,(4)$ & $160\,(18)$ \\
Greedy pruning, neurons removed with test $\geq 0.95$ & $177\,(5)$ & $117\,(37)$ \\
\bottomrule
\end{tabular}
\end{center}
\end{table}

\begin{table}[!ht]
\caption{Deletion tolerance under random and norm-ranked deletion (modular addition, $p=31$, three seeds). Norm-ranked deletion removes neurons in decreasing order of outgoing norm.}
\label{tab:targeted}
\begin{center}
\small
\begin{tabular}{lccc}
\toprule
Deletion procedure & Reference & Credit equalization ($\beta{=}1$) & DFC \\
\midrule
Random        & $0.674$ & $\mathbf{0.770}$ & $0.725$ \\
Norm-ranked   & $0.332$ & $\mathbf{0.581}$ & $0.518$ \\
\bottomrule
\end{tabular}
\end{center}
\end{table}

\begin{table}[!ht]
\caption{Greedy neuron deletion on modular addition ($p=31$, three seeds). At each step the procedure removes the neuron whose removal lowers test accuracy the most. The last column gives the mean number of neurons removed before test accuracy falls below $0.5$. Standard deviations of circuit participation are in parentheses.}
\label{tab:plane}
\begin{center}
\small
\begin{tabular}{lccc}
\toprule
Condition & Test & Circuit PR & Neurons removed \\
\midrule
Reference & $0.998$ & $0.595\,(0.023)$ & $23.3$ \\
Credit equalization, $\beta{=}0.25$ & $0.997$ & $0.940\,(0.017)$ & $43.0$ \\
Credit equalization, $\beta{=}0.5$ & $0.986$ & $0.975\,(0.002)$ & $59.0$ \\
Credit equalization, $\beta{=}1$ & $0.976$ & $0.992\,(0.002)$ & $60.7$ \\
\bottomrule
\end{tabular}
\end{center}
\end{table}

\subsection{Minimal circuits}
\label{app:minimal}
We prune the first hidden layer of each trained multilayer perceptron, or the MLP of the transformer, and record the test-accuracy curve along the pruning order and the number of neurons kept just before it first falls below a threshold. Greedy pruning removes at each step the neuron whose removal raises the training cross-entropy least, recomputed after every removal, so the order uses only training pairs and the held-out pairs only set where the curve is read. Magnitude pruning removes neurons in increasing order of outgoing norm. All networks are trained on modular addition at $p = 31$ with five fresh seeds. The network with a decayed per-neuron gain multiplies each first-layer pre-activation by a trainable gain with its own bias, both decayed like the weights, and the transformer is the wide-MLP configuration of Appendix~\ref{app:transformer}. Greedy pruning to test accuracy $0.95$ changes the mean trigonometric fit of Appendix~\ref{app:algo} by at most $0.11$ in the multilayer perceptrons, for example from $0.66$ to $0.55$ for the reference and from $0.90$ to $0.81$ under equalization at width $2048$, so their pruned circuits keep the Fourier structure of the full networks (Table~\ref{tab:minimalcheck}). In the transformer the fit falls from $0.99$ to $0.44$ in both conditions, which lies in the range $0.42$ to $0.68$ of the full networks of the grokking rules in Appendix~\ref{app:algo}.

\begin{table}[!ht]
\caption{Number of neurons that pruning keeps at test accuracy $0.90$ or $0.95$ (modular addition, $p=31$, five seeds; range over seeds in brackets).}
\label{tab:minimal}
\begin{center}
\footnotesize
\setlength{\tabcolsep}{4pt}
\begin{tabular}{llccccc}
\toprule
Network & Width & Circuit PR & Greedy, $0.90$ & Greedy, $0.95$ & Magnitude, $0.95$ \\
\midrule
Adam with L2, reference & $256$ & $0.594$ & $59$ & $70$ $[63, 81]$ & $88$ \\
Adam with L2, equalized & $256$ & $0.993$ & $96$ & $155$ $[100, 230]$ & $215$ \\
Decayed per-neuron gain & $256$ & $0.096$ & $23$ & $24$ $[22, 25]$ & $25$ \\
AdamW, reference & $256$ & $0.931$ & $61$ & $77$ $[67, 96]$ & $142$ \\
AdamW, equalized & $256$ & $0.970$ & $61$ & $115$ $[71, 171]$ & $217$ \\
Adam-mini with L2 & $256$ & $0.586$ & $61$ & $72$ $[55, 90]$ & $89$ \\
\midrule
Adam with L2, reference & $2048$ & $0.168$ & $58$ & $73$ $[33, 98]$ & $143$ \\
Adam with L2, equalized & $2048$ & $0.999$ & $132$ & $171$ $[138, 229]$ & $1692$ \\
Decayed per-neuron gain & $2048$ & $0.009$ & $16$ & $18$ $[14, 22]$ & $18$ \\
\midrule
Transformer, reference & $4096$ & $0.609$ & $111$ & $140$ $[89, 171]$ & $258$ \\
Transformer, equalized & $4096$ & $0.994$ & $395$ & $462$ $[416, 536]$ & $1120$ \\
\bottomrule
\end{tabular}
\end{center}
\end{table}

At width $2048$ the greedy circuit at threshold $0.95$ under equalization is $7.1$ to $13.5$ times that of the gain network. The order gain network, reference, equalization holds in every seed at thresholds $0.90$ and $0.95$ and in both pruning orders. With the architecture fixed, equalization enlarges the greedy circuit $2.4$ times and the magnitude circuit $12$ times. In the distributed networks greedy pruning finds far smaller circuits than magnitude pruning, because nearly equal outgoing norms do not indicate which neurons the computation needs. Participation and minimal circuit size measure different things: the AdamW reference has nearly equal outgoing norms but prunes to about as few neurons as the Adam reference.

\paragraph{Starting accuracy and stopping point.} The networks reach different test accuracies before pruning, and a network that starts closer to the threshold crosses it after fewer removals. We therefore also stop one or two points below each network's own test accuracy (Table~\ref{tab:minimalcheck}). At width $2048$ the reference and equalization both start at $0.998$, and the order gain network, reference, equalization holds in every seed at both drops and in both pruning orders; two points below, the greedy circuit under equalization is $2.6$ times that of the reference. At width $256$, where equalization starts lower, it is $1.8$ times that of the reference and larger in every seed. The gain network starts lower than the reference, which would enlarge its circuit rather than shrink it. Taking the smallest size at which the curve is still at or above $0.95$, instead of the first crossing, changes the greedy means at width $2048$ from $73$, $171$ and $18$ to $71$, $162$ and $17$.

\begin{table}[!ht]
\caption{Pruning relative to each network's own test accuracy (same networks as Table~\ref{tab:minimal}, means over five seeds). Test is the accuracy before pruning; the next two columns give the neurons that greedy pruning keeps one and two points below it. Fit is the trigonometric fit of Appendix~\ref{app:algo} for the full network and for the greedy circuit at test accuracy $0.95$.}
\label{tab:minimalcheck}
\begin{center}
\footnotesize
\setlength{\tabcolsep}{4pt}
\begin{tabular}{llccccc}
\toprule
Network & Width & Test & Greedy, $-0.01$ & Greedy, $-0.02$ & Fit, full & Fit, pruned \\
\midrule
Adam with L2, reference & $256$ & $0.998$ & $99$ & $82$ & $0.54$ & $0.47$ \\
Adam with L2, equalized & $256$ & $0.980$ & $199$ & $146$ & $0.47$ & $0.44$ \\
Decayed per-neuron gain & $256$ & $0.975$ & $25$ & $25$ & $0.49$ & $0.48$ \\
AdamW, reference & $256$ & $0.992$ & $109$ & $95$ & $0.36$ & $0.42$ \\
AdamW, equalized & $256$ & $0.967$ & $116$ & $88$ & $0.43$ & $0.41$ \\
Adam-mini with L2 & $256$ & $0.995$ & $92$ & $83$ & $0.58$ & $0.51$ \\
\midrule
Adam with L2, reference & $2048$ & $0.998$ & $128$ & $97$ & $0.66$ & $0.55$ \\
Adam with L2, equalized & $2048$ & $0.998$ & $342$ & $252$ & $0.90$ & $0.81$ \\
Decayed per-neuron gain & $2048$ & $0.972$ & $18$ & $18$ & $0.75$ & $0.75$ \\
\midrule
Transformer, reference & $4096$ & $1.000$ & $364$ & $249$ & $0.99$ & $0.44$ \\
Transformer, equalized & $4096$ & $1.000$ & $1208$ & $867$ & $0.99$ & $0.44$ \\
\bottomrule
\end{tabular}
\end{center}
\end{table}

\subsection{Matching participation through outgoing-weight constraints}
\label{app:suff}
Does the deletion advantage depend on how the distributed circuit was trained, or only on the resulting distribution of outgoing norms? We compare two ways of reaching the same circuit participation. Credit equalization changes the gradients. An \emph{outgoing-norm constraint} leaves the gradients unchanged and instead rescales the outgoing weights of the first hidden layer after every update to a prescribed norm profile with participation $0.60$, $0.80$ or $0.99$; the profile has two norm levels, each shared by half of the neurons. All conditions use five seeds (Table~\ref{tab:suff}).

\paragraph{Deletion.} At participation $0.99$, credit equalization and the outgoing-norm constraint give the same random-deletion tolerance, $0.777$; at participation $0.60$, the constrained network and the reference differ by $0.02$. Greedy deletion removes $66.7\,(11.1)$ neurons under equalization and $68.7\,(10.8)$ under the constraint before accuracy falls below $0.5$. Deletion tolerance thus follows the distribution of outgoing norms in both ways of producing it.

\paragraph{Adaptation.} We also fine-tune these networks from addition to subtraction and measure the epochs to test accuracy $0.9$, evaluated every $20$ epochs. Within the constrained networks, adaptation takes $740$, $1228$ and $1480$ epochs at participation $0.60$, $0.80$ and $0.99$; the equalized network takes $1608$ epochs. Adaptation slows with participation under both interventions. At participation $0.60$, the reference takes $464$ epochs and the constrained network $740$: the reference has $87$ neurons with an outgoing norm below one tenth of the layer mean, which the new task can recruit, whereas the constraint holds every neuron at one of two norms and leaves none. The difference points to the low-norm neurons, which a related task can recruit.

\begin{table}[!ht]
\caption{Credit equalization compared with an outgoing-norm constraint of matched participation (modular addition, $p=31$, five seeds; standard deviations in parentheses). Random del is the random-deletion tolerance; Adapt is the number of fine-tuning epochs to test accuracy $0.9$ on subtraction.}
\label{tab:suff}
\begin{center}
\small
\begin{tabular}{lcccc}
\toprule
Condition & Test & Circuit PR & Random del & Adapt \\
\midrule
Reference            & $0.995\,(0.007)$ & $0.597\,(0.020)$ & $0.669$ & $464$ \\
Credit equalization ($\beta{=}1$)   & $0.976\,(0.015)$ & $0.993\,(0.002)$ & $0.777$ & $1608$ \\
\midrule
Outgoing-norm constraint, $0.60$   & $0.987\,(0.016)$ & $0.600\,(0.000)$ & $0.647$ & $740$ \\
Outgoing-norm constraint, $0.80$   & $0.977\,(0.008)$ & $0.800\,(0.000)$ & $0.675$ & $1228$ \\
Outgoing-norm constraint, $0.99$   & $0.976\,(0.012)$ & $0.990\,(0.000)$ & $0.777$ & $1480$ \\
\bottomrule
\end{tabular}
\end{center}
\end{table}

\subsection{Adaptation to a new operation}
\label{app:adapt}
Networks first grok addition at $p = 31$ at each $\beta$. We then fine-tune the whole network on the new operation with the reference rule, backpropagation with Adam at learning rate $10^{-3}$ and a fresh optimizer state, in every condition, on the new operation's own split with $40\%$ of the pairs held out, and record the number of epochs until test accuracy on the new operation reaches $0.9$, evaluated every $100$ epochs (Table~\ref{tab:adapt}).

\paragraph{Transfer to subtraction.} Adaptation time increases with the participation of the source circuit at every step, from $640$ epochs at $\beta = 0$ to $1540$ at $\beta = 1$; training from scratch takes $1410$ epochs. The seed ranges at the two endpoints do not overlap: $400$ to $800$ epochs at $\beta = 0$ against $1200$ to $1800$ at $\beta = 1$. An independent ten-seed run, evaluated every $20$ epochs, gives $690\,(164)$ and $1550\,(191)$ epochs. Pretraining on addition helps from the concentrated circuit and not from the distributed one.

\paragraph{The role of low-norm neurons.} We count the neurons whose outgoing norm is below one tenth of the layer mean. The reference has about $85$ such neurons. Our hypothesis is that subtraction can recruit them with little disturbance to the addition circuit. Already at $\beta = 0.25$ the count is zero, and adaptation time rises from $640$ to $940$ epochs. It keeps rising to $1540$ epochs at $\beta = 1$ as the circuit spreads further, which is consistent with the new operation having to modify a larger share of a distributed representation. In continual learning, inactive units signal a loss of plasticity that resetting them restores \citep{sokar2023dormant,dohare2024loss}; here the low-norm neurons of the reference circuit coincide with faster adaptation, which suggests that unused capacity can serve a related task when the rest of the circuit stays intact.

\paragraph{Transfer after switching the learning rule.} We train reference networks on addition for $20{,}000$ epochs, then copy each into two branches that continue for $30{,}000$ epochs with fresh Adam states, using either the reference rule or full equalization. Both branches are then fine-tuned identically on subtraction, with evaluation every $20$ epochs. Across five seeds, equalized branches reach $0.9$ test accuracy in $1052\,(273)$ epochs versus $608\,(160)$ for reference continuations; the mean paired difference is $444\,(131)$ epochs. Source accuracies are $0.987$ and $0.996$, respectively. Equalized versus reference continuations have outgoing-weight participation of $0.99$ versus $0.60$ and ablation participation of $0.72$ versus $0.40$. The participation measures show little further change between $10{,}000$ and $30{,}000$ continuation epochs. Parentheses give standard deviations.

\paragraph{Fine-tuning with the source rule.} The transfer runs above fine-tune with the reference rule. To check that the ordering does not depend on the fine-tuning rule, we fine-tune each source network with both rules, evaluated every $20$ epochs (Table~\ref{tab:ownrule}). Over five seeds, the concentrated network reaches $0.9$ on subtraction after $580\,(151)$ epochs with the reference rule and $1000\,(196)$ with full equalization; the distributed network takes $1496\,(195)$ and $1856\,(95)$. The distributed source is slower under both fine-tuning rules and in every seed, by a factor of $2.6$ and $1.9$. The slower adaptation of the distributed circuit does not depend on the fine-tuning rule.

\begin{table}[!ht]
\caption{Fine-tuning from addition to subtraction with the reference rule or with full credit equalization, for a concentrated and a distributed source network (modular addition, $p=31$, five seeds; standard deviations in parentheses). Adaptation time is the number of epochs to test accuracy $0.9$, evaluated every $20$ epochs.}
\label{tab:ownrule}
\begin{center}
\small
\begin{tabular}{llccc}
\toprule
Source network & Fine-tuning rule & Source test & Epochs to $0.9$ & Circuit PR after \\
\midrule
Concentrated ($\beta{=}0$) & reference & $0.992$ & $580\,(151)$ & $0.62$ \\
Concentrated ($\beta{=}0$) & equalization & $0.992$ & $1000\,(196)$ & $0.99$ \\
Distributed ($\beta{=}1$) & reference & $0.969$ & $1496\,(195)$ & $0.60$ \\
Distributed ($\beta{=}1$) & equalization & $0.969$ & $1856\,(95)$ & $0.99$ \\
\bottomrule
\end{tabular}
\end{center}
\end{table}

\paragraph{Nearly unused neurons and the kind of target.} We pretrain addition networks at $\beta = 0$ and $\beta = 1$ for $20{,}000$ epochs and fine-tune them as above on five targets, evaluated every $20$ epochs (Table~\ref{tab:spare}, five seeds). A third source is the concentrated network with its nearly unused neurons removed. Their incoming weights, bias and outgoing weights are set to zero, which removes $75$ to $96$ neurons and leaves test accuracy on addition at $0.996$; they stay at zero during fine-tuning, since a ReLU unit with zero incoming weights and bias receives no gradient. Removing them slows adaptation by $552\,(217)$ epochs on subtraction and $492\,(260)$ epochs on $(b - a) \bmod p$, in every seed, which is $67\%$ and $54\%$ of the difference to the distributed network. Two targets keep the hidden features of addition and change only the readout, $(a + b + 7) \bmod p$ and $\pi((a + b) \bmod p)$ for a fixed random permutation $\pi$. Both are learned from every source $2.3$ to $6.6$ times faster than from scratch, and the concentrated network has no advantage.

\begin{table}[!ht]
\caption{Adaptation from addition to five targets, in epochs to test accuracy $0.9$ (modular addition source, $p=31$, five seeds; standard deviations in parentheses). The third column is the concentrated network with its nearly unused neurons removed.}
\label{tab:spare}
\begin{center}
\small
\setlength{\tabcolsep}{4pt}
\begin{tabular}{lcccc}
\toprule
Target & Concentrated & Nearly unused removed & Distributed & From scratch \\
\midrule
$(a - b) \bmod p$ & $652\,(108)$ & $1204\,(303)$ & $1480\,(382)$ & $1300\,(83)$ \\
$(b - a) \bmod p$ & $528\,(72)$ & $1020\,(296)$ & $1440\,(303)$ & $1304\,(46)$ \\
$(a + b + 7) \bmod p$ & $204\,(60)$ & $204\,(69)$ & $224\,(39)$ & $1356\,(97)$ \\
$\pi((a + b) \bmod p)$ & $560\,(115)$ & $452\,(95)$ & $464\,(83)$ & $1308\,(111)$ \\
$(a \cdot b) \bmod p$ & $1940\,(124)$ & $2324\,(224)$ & $1988\,(202)$ & $1148\,(52)$ \\
\bottomrule
\end{tabular}
\end{center}
\end{table}

\paragraph{Transfer follows the circuit.} We also fine-tune the four networks of Table~\ref{tab:decay} on subtraction and on $(b - a) \bmod p$, five seeds each. The two networks with distributed circuits, produced by credit equalization or by per-neuron decay alone, take $1572\,(189)$ and $1408\,(141)$ epochs on subtraction and $1560\,(279)$ and $1452\,(74)$ on $(b - a) \bmod p$. The two networks with concentrated circuits, the reference and equalization with the reference's ratio of gradient to decay, take $616\,(117)$ and $536\,(86)$ on subtraction and $528\,(77)$ and $512\,(92)$ on $(b - a) \bmod p$. On both targets every run from a distributed circuit is slower than every run from a concentrated one. Across these four training conditions, adaptation time thus follows the circuit rather than the rule that produced it.

\paragraph{Transfer to multiplication.} The same pretrained networks take $1800$ to $2020$ epochs to reach the threshold on multiplication, against $1200$ from scratch. Multiplication does not share the operand frequencies of addition, and pretraining slows adaptation at every $\beta$. In our experiments the advantage of the concentrated circuit appears only for the target task that shares the operand frequencies of the source task.

\paragraph{Fine-tuning learning rate.} At one third and at three times the reference learning rate, adaptation time still increases with $\beta$, and the fully equalized start takes $1.9$ to $2.5$ times as long as the concentrated start (Table~\ref{tab:ftlr}). The reverse transfer, from subtraction to addition, shows the same ordering: $740$ epochs at $\beta = 0$ and $1600$ at $\beta = 1$.

\begin{table}[!ht]
\caption{Fine-tuning from addition to a new operation, with standard deviations in parentheses. Subtraction results for $\beta = 0, 0.25, 1$ and from scratch use ten seeds; the intermediate settings and all multiplication results use five. Adaptation time is the number of epochs to test accuracy $0.9$. Low-norm neurons have an outgoing norm below one tenth of the layer mean.}
\label{tab:adapt}
\begin{center}
\footnotesize
\begin{tabular}{lccccc}
\toprule
Condition & Circuit PR & Low-norm neurons & Source test & \multicolumn{2}{c}{Epochs to $0.9$ on} \\
 & & & & subtraction & multiplication \\
\midrule
$\beta=0$ (reference) & $0.593\,(0.034)$ & $85$ & $0.995$ & $640\,(128)$  & $1800\,(141)$ \\
$\beta=0.25$         & $0.933\,(0.020)$ & $0$  & $0.989$ & $940\,(191)$  & $1940\,(233)$ \\
$\beta=0.5$          & $0.972\,(0.006)$ & $0$  & $0.990$ & $1200\,(89)$  & $1980\,(248)$ \\
$\beta=0.75$         & $0.983\,(0.003)$ & $0$  & $0.981$ & $1300\,(141)$ & $2020\,(133)$ \\
$\beta=1$            & $0.991\,(0.002)$ & $0$  & $0.972$ & $1540\,(180)$ & $1940\,(136)$ \\
From scratch         & ---              & ---  & ---     & $1410\,(137)$ & $1200\,(63)$ \\
\bottomrule
\end{tabular}
\end{center}
\end{table}

\begin{table}[!ht]
\caption{Adaptation time from addition to subtraction across an order of magnitude in the fine-tuning learning rate, in epochs to test accuracy $0.9$ (standard deviations in parentheses). The outer columns use three seeds; the middle column repeats Table~\ref{tab:adapt} (ten seeds for the endpoints, five for $\beta = 0.5$).}
\label{tab:ftlr}
\begin{center}
\small
\begin{tabular}{lccc}
\toprule
Start & $3\times 10^{-4}$ & $10^{-3}$ & $3\times 10^{-3}$ \\
\midrule
$\beta=0$ (reference) & $1933\,(525)$ & $640\,(128)$  & $267\,(47)$ \\
$\beta=0.5$          & $3067\,(205)$ & $1200\,(89)$  & $533\,(47)$ \\
$\beta=1$            & $3733\,(262)$ & $1540\,(180)$ & $667\,(94)$ \\
\midrule
ratio $\beta{=}1$ to $\beta{=}0$ & $1.9$ & $2.4$ & $2.5$ \\
\bottomrule
\end{tabular}
\end{center}
\end{table}

\section{Other tasks and architectures}
\label{app:scope}
The main experiments use modular addition and a multilayer perceptron. Here we repeat the comparison between the reference and full credit equalization on other operations, tasks and architectures (Table~\ref{tab:scope}). The intervention raises participation where the reference circuit is concentrated, with one exception, layer normalization, and it changes little where the reference circuit is already distributed.

\begin{table}[!ht]
\caption{Circuit participation of the reference ($\beta{=}0$) and under full credit equalization ($\beta{=}1$) across the tested settings. Test accuracies and seed counts are given in the cited subsections. Sparse parity and MNIST show no grokking delay. The last three settings are unchanged: two start from a distributed circuit, and with layer normalization the reference is concentrated.}
\label{tab:scope}
\begin{center}
\footnotesize
\setlength{\tabcolsep}{5pt}
\begin{tabular}{llccl}
\toprule
Setting & Circuit measured & $\beta{=}0$ & $\beta{=}1$ & Details \\
\midrule
Addition, subtraction, multiplication & first hidden layer & $0.43$--$0.60$ & $0.99$ & App.~\ref{app:ops} \\
Sparse parity           & first hidden layer & $0.15$ & $0.96$ & App.~\ref{app:parity} \\
$S_5$ composition       & first hidden layer & $0.70$ & $0.97$ & App.~\ref{app:groups} \\
Transformer, $p=31$, wide MLP & MLP block & $0.61$ & $0.99$ & App.~\ref{app:transformer} \\
Batch normalization & first hidden layer & $0.81$ & $0.97$ & App.~\ref{app:normlayers} \\
\midrule
Transformer, $p=113$, canonical & MLP block & $0.96$ & $0.96$ & App.~\ref{app:transformer} \\
MNIST                   & first hidden layer & $0.99$ & $0.99$ & App.~\ref{app:mnist} \\
Layer normalization & first hidden layer & $0.36$ & $0.37$ & App.~\ref{app:normlayers} \\
\bottomrule
\end{tabular}
\end{center}
\end{table}

\subsection{Subtraction and multiplication}
\label{app:ops}
Table~\ref{tab:ops} repeats the comparison on subtraction and multiplication at $p = 31$, with DFC for comparison. Equalization raises participation to $0.99$ on both operations while test accuracy stays at or above $0.994$, and random-deletion tolerance increases as on addition. The largest change is on multiplication, where the reference circuit is the most concentrated of the three, with participation $0.43$.

\begin{table}[!ht]
\caption{Circuit participation and random-deletion tolerance on the three modular operations ($p=31$, three seeds; standard deviations in parentheses). All three operations use the same random split with $40\%$ of the pairs held out.}
\label{tab:ops}
\begin{center}
\small
\begin{tabular}{llccc}
\toprule
Operation & Rule & Test & Circuit PR & Random del \\
\midrule
Addition       & reference & $0.998\,(0.001)$ & $0.60\,(0.02)$ & $0.671\,(0.016)$ \\
               & equalization ($\beta{=}1$) & $0.976\,(0.011)$ & $0.99\,(0.00)$ & $0.771\,(0.019)$ \\
               & DFC & $1.000\,(0.000)$ & $0.92\,(0.00)$ & $0.717\,(0.014)$ \\
\midrule
Subtraction    & reference & $0.999\,(0.001)$ & $0.59\,(0.04)$ & $0.680\,(0.026)$ \\
               & equalization ($\beta{=}1$) & $0.994\,(0.005)$ & $0.99\,(0.00)$ & $0.785\,(0.014)$ \\
               & DFC & $1.000\,(0.000)$ & $0.93\,(0.01)$ & $0.735\,(0.010)$ \\
\midrule
Multiplication & reference & $0.999\,(0.001)$ & $0.43\,(0.07)$ & $0.666\,(0.035)$ \\
               & equalization ($\beta{=}1$) & $0.999\,(0.001)$ & $0.99\,(0.00)$ & $0.867\,(0.012)$ \\
               & DFC & $1.000\,(0.000)$ & $0.86\,(0.02)$ & $0.764\,(0.010)$ \\
\bottomrule
\end{tabular}
\end{center}
\end{table}

\subsection{Network width}
\label{app:arch}
Does the difference between the two circuits depend on how many neurons are available? We vary the hidden width from $32$ to $2048$ with the task, the budget and the comparison between $\beta = 0$ and $\beta = 1$ fixed (Table~\ref{tab:widths}). At width $32$ the effective counts are $30$ for the reference and $31$ under equalization. At width $2048$ they are $332$ and $2046$. The reference uses a decreasing fraction of the layer as the width grows, from participation $0.937$ to $0.162$, while equalization stays near one. Test accuracy is at least $0.96$ throughout. Normalized participation and effective count answer different questions and can move in opposite directions: from width $128$ to $512$ the reference's participation falls from $0.791$ to $0.400$ while its effective count grows from $101$ to $205$. A wider reference network uses more neurons in absolute terms and a smaller share of the layer.

\begin{table}[!ht]
\caption{Width dependence of first-hidden-layer circuit participation and effective neuron count $n \cdot \mathrm{PR}$ at $\beta = 0$ and $\beta = 1$ (modular addition, $p=31$, three seeds, $15{,}000$ epochs).}
\label{tab:widths}
\begin{center}
\small
\begin{tabular}{rcccc}
\toprule
& \multicolumn{2}{c}{Circuit PR} & \multicolumn{2}{c}{$n \cdot \mathrm{PR}$} \\
Width & $\beta=0$ & $\beta=1$ & $\beta=0$ & $\beta=1$ \\
\midrule
$32$   & $0.937$ & $0.980$ & $30$  & $31$   \\
$64$   & $0.915$ & $0.987$ & $59$  & $63$   \\
$128$  & $0.791$ & $0.991$ & $101$ & $127$  \\
$512$  & $0.400$ & $0.994$ & $205$ & $509$  \\
$1024$ & $0.262$ & $0.994$ & $268$ & $1018$ \\
$2048$ & $0.162$ & $0.999$ & $332$ & $2046$ \\
\bottomrule
\end{tabular}
\end{center}
\end{table}

\subsection{Sparse parity}
\label{app:parity}
We test credit equalization on sparse parity, a task used to study how networks find a few relevant features after a long plateau \citep{barak2022hidden}. The input is a vector of $20$ bits, each $0$ or $1$; the label is $1$ if an odd number of three fixed bits are set and $0$ otherwise, so the remaining $17$ bits are distractors that the network has to ignore. We draw $4000$ or $600$ inputs at random from the $2^{20}$ possibilities and assign each to the training or the test set with probability one half, which gives about $2000$ and $300$ training examples. The runs generalize without delay under full-batch training, so this experiment tests the redistribution outside the grokking regime. Equalization raises circuit participation from $0.15$ to $0.96$ on the larger dataset and from $0.08$ to $0.89$ on the smaller one. Test accuracy is $1.000$ for both conditions on the larger dataset and changes from $1.000$ to $0.988$ on the smaller one, and random-deletion tolerance improves at both sizes (Table~\ref{tab:parity}). The redistribution is thus not restricted to modular arithmetic or to delayed generalization.

\begin{table}[!ht]
\caption{Sparse parity with approximately $2000$ and $300$ training examples (three seeds; standard deviations in parentheses).}
\label{tab:parity}
\begin{center}
\small
\begin{tabular}{llccc}
\toprule
Train size & Rule & Test & Circuit PR & Random del \\
\midrule
$\sim 2000$ & reference & $1.000\,(0.000)$ & $0.15\,(0.06)$ & $0.865\,(0.051)$ \\
            & equalization ($\beta{=}1$) & $1.000\,(0.000)$ & $0.96\,(0.03)$ & $0.955\,(0.002)$ \\
\midrule
$\sim 300$  & reference & $1.000\,(0.000)$ & $0.08\,(0.02)$ & $0.825\,(0.017)$ \\
            & equalization ($\beta{=}1$) & $0.988\,(0.017)$ & $0.89\,(0.02)$ & $0.908\,(0.060)$ \\
\bottomrule
\end{tabular}
\end{center}
\end{table}

\subsection{Permutation composition}
\label{app:groups}
We test whether credit equalization redistributes circuits on a noncommutative operation, the composition of permutations, which \citet{chughtai2023toy} use to study the circuits of grokked networks. The group $S_5$ contains the $120$ permutations of five objects. The input is a pair $(g, h)$ of permutations, and the target is their composition $g \circ h$, the permutation obtained by applying $h$ first and then $g$. We use one-hot inputs and train for $30{,}000$ epochs on half of the $120 \times 120$ input pairs, with the rest held out. Both the reference and the equalized network grok under the same definition as for modular arithmetic; the delay grows when fewer training pairs are available. Full equalization raises circuit participation from $0.699\,(0.078)$ to $0.973\,(0.005)$ while both conditions reach test accuracy $1.000$. Random-deletion tolerance increases from $0.485$ to $0.633$ and norm-ranked tolerance from $0.149$ to $0.468$ (Table~\ref{tab:groups}). Reducing the training fraction to $40\%$ reproduces the pattern. Credit and circuit participation again agree closely.

\begin{table}[!ht]
\caption{Composition in $S_5$. The upper block uses ten seeds with half the pairs held out, the lower block five seeds with $60\%$ held out; parentheses show standard deviations. Delay is the number of epochs between reaching training accuracy $1.0$ and test accuracy $0.95$.}
\label{tab:groups}
\begin{center}
\small
\begin{tabular}{lcccccc}
\toprule
Rule & Test & Credit PR & Circuit PR & Random del & Norm-ranked del & Delay \\
\midrule
\multicolumn{7}{l}{\emph{Half the pairs held out, ten seeds}} \\
Reference & $1.000$ & $0.689$ & $0.699\,(0.078)$ & $0.485$ & $0.149$ & $500$ \\
Equalization, $\beta{=}0.5$ & $1.000$ & $0.970$ & $0.942\,(0.010)$ & $0.585$ & $0.322$ & $550$ \\
Equalization, $\beta{=}1$ & $1.000$ & $1.000$ & $0.973\,(0.005)$ & $0.633$ & $0.468$ & $900$ \\
\midrule
\multicolumn{7}{l}{\emph{$60\%$ held out, five seeds}} \\
Reference & $1.000$ & $0.657$ & $0.662\,(0.038)$ & $0.409$ & $0.129$ & $1350$ \\
Equalization, $\beta{=}0.5$ & $1.000$ & $0.951$ & $0.907\,(0.016)$ & $0.525$ & $0.260$ & $1650$ \\
Equalization, $\beta{=}1$ & $1.000$ & $1.000$ & $0.970\,(0.002)$ & $0.597$ & $0.410$ & $1750$ \\
\bottomrule
\end{tabular}
\end{center}
\end{table}

\subsection{Transformers}
\label{app:transformer}
We test whether credit equalization changes the participation of a transformer's MLP block. We follow the modular-addition transformer of \citet{nanda2023progress}: inputs have the form ``$a\ b\ =$'', and the canonical configuration has one block, $p = 113$, $d_{\text{model}} = 128$, one attention head per $32$ channels, a ReLU MLP of width $4 d_{\text{model}}$ and weight decay $1.0$, trained with AdamW ($\beta_1 = 0.9$, $\beta_2 = 0.98$) at learning rate $10^{-3}$ on $30\%$ of the operand pairs for $40{,}000$ epochs. Credit equalization is applied to every linear map.

In the canonical model the reference MLP is already distributed: its participation is $0.955$, and full equalization leaves it at $0.955$ with test accuracy $1.000$ in both conditions (Table~\ref{tab:transformer}). A second modulus and a doubled model width give the same picture, with MLP participation changing by at most $0.015$. Applying the intervention only to the MLP changes participation by less than the variation across seeds. The intervention thus does not increase MLP participation where the reference is already distributed.

A wider MLP gives a concentrated reference. With $4096$ MLP neurons at $p = 31$, reference participation is $0.610\,(0.048)$, and full equalization raises it to $0.993\,(0.003)$ over five seeds, with the seed ranges not overlapping. Norm-ranked deletion tolerance rises from $0.580$ to $0.941$, test accuracy stays at $1.000$ and the margin is unchanged to two decimal places.

In the wide configuration, equalization acts on the MLP's incoming weights only, training uses $60\%$ of the operand pairs for $60{,}000$ epochs with a warmup followed by cosine decay, and AdamW decays every parameter, including the layer-normalization gains and shifts, as in the canonical setting. After about $20{,}000$ epochs the square roots of Adam's second-moment estimates for the MLP's incoming weights fall below $\epsilon = 10^{-8}$, to $10^{-11}$ to $5 \cdot 10^{-11}$ at the end, so Adam's step is approximately $\hat{\vm}/\epsilon$ and $\epsilon$ acts as the non-gradient term of the update. On five fresh seeds, the reference has participation $0.609$ and full equalization $0.994$. Dividing $\epsilon$ by each neuron's equalization factor, with the gradient unchanged, gives $0.981$. Equalizing the gradient and multiplying $\epsilon$ by the same factor, which keeps each neuron's ratio of gradient to $\epsilon$, gives $0.818$, so this control removes about half of the effect. With decay only on weight matrices, as in most current training code, $\epsilon$ no longer dominates. The reference is then broadly distributed, with participation $0.82$ over ten seeds, and equalization raises it to $0.91$; one of the ten equalized runs did not generalize.

Credit equalization thus has a large effect in the transformer whose reference MLP is concentrated and a small one where the reference is already distributed.

\begin{table}[!ht]
\caption{Modular-addition transformer results: test accuracy and MLP circuit participation (standard deviations in parentheses). Credit equalization is applied to every linear map; $\beta{=}0$ is the reference. The wide-MLP result is described in the text.}
\label{tab:transformer}
\begin{center}
\small
\begin{tabular}{lcc}
\toprule
$\beta$ & Test & MLP circuit PR \\
\midrule
\multicolumn{3}{l}{\emph{Canonical: one block, $p=113$, $d_{\text{model}}=128$, six seeds}} \\
$0.0$ & $1.000$ & $0.955\,(0.01)$ \\
$0.5$ & $1.000$ & $0.953\,(0.01)$ \\
$1.0$ & $1.000$ & $0.955\,(0.01)$ \\
\midrule
\multicolumn{3}{l}{\emph{Second modulus: $p=97$, three seeds, $70{,}000$ epochs}} \\
$0.0$ & $1.000$ & $0.959\,(0.01)$ \\
$1.0$ & $1.000$ & $0.951\,(0.02)$ \\
\midrule
\multicolumn{3}{l}{\emph{Twice the width: $d_{\text{model}}=256$, three seeds}} \\
$0.0$ & $1.000$ & $0.902\,(0.04)$ \\
$1.0$ & $1.000$ & $0.917\,(0.01)$ \\
\bottomrule
\end{tabular}
\end{center}
\end{table}

\subsection{MNIST}
\label{app:mnist}
We use the setting in which image-classification grokking was reported \citep{liu2022omnigrok}: $1000$ training images and a network with two hidden layers of width $200$, at the standard and at a larger initialization scale. Training and test accuracy rise together, so these runs show no grokking delay. The reference already has circuit participation $0.992\,(0.002)$, and full credit equalization changes it by less than the variation across seeds. As in the canonical transformer, the reference leaves little room for a further increase in participation.

\subsection{Normalization layers}
\label{app:normlayers}
We insert batch normalization \citep{ioffe2015batchnorm} or layer normalization \citep{ba2016layernorm} after each hidden linear map of the multilayer perceptron, before the ReLU, and otherwise train as in Section~\ref{sec:setup} for $20{,}000$ epochs over five seeds. Measurements use evaluation mode, with the running statistics for batch normalization. Batch normalization, which normalizes each neuron's pre-activation separately, partly distributes the circuit, to participation $0.808$ against $0.611$ without normalization, with $42$ instead of $81$ nearly unused neurons, and equalization raises it to $0.970$ (Table~\ref{tab:normlayers}). Layer normalization, which normalizes across the neurons of a layer, concentrates the circuit further, to participation $0.356$ with $164$ nearly unused neurons, and equalizing the first layer's gradients leaves it at $0.369$; on the five fresh seeds of Table~\ref{tab:pathways} the two values are $0.366$ and $0.364$. Test accuracy is $1.000$ with either normalization layer. Layer normalization concentrates the circuit through its trainable per-neuron gain, not through the normalization. With the normalization layer on the first hidden layer only, a trainable gain and bias without any normalization give the same participation as layer normalization, $0.099$ against $0.094$ with $231$ nearly unused neurons each, and RMS normalization gives $0.094$. Layer normalization without a trainable gain gives $0.53$ to $0.58$, and batch normalization without affine parameters gives $0.996$ with no intervention. Under layer normalization a neuron has two decayed scales of its own, the gain and the outgoing weights, and equalizing the incoming gradients reaches neither. Rescaling the decay of both by the inverse of each neuron's equalization factor distributes the circuit in five fresh seeds, with participation $0.997$ and no nearly unused neurons when layer normalization follows both hidden maps, and $1.000$ at test accuracy $0.94$ to $1.00$ when it follows the first. Equalizing the gradients of the incoming rows, the gain and the outgoing weights does the same, with $0.996$ and $1.000$, while equalizing them with each neuron's ratio of gradient to decay held fixed leaves participation at $0.343$ and $0.129$. In an earlier set of five seeds the activation-weighted participation follows, at $0.95$ and $0.98$ with the decay rescaled and $0.31$ and $0.12$ with the ratio held. With layer normalization on the first hidden layer, the circuit keeps about $26$ neurons at widths $64$, $256$ and $1024$, against $61$, $175$ and $305$ without normalization.

\begin{table}[!ht]
\caption{Normalization layers after each hidden linear map (modular addition, $p=31$, five seeds; standard deviations in parentheses). Columns as in Table~\ref{tab:decay}.}
\label{tab:normlayers}
\begin{center}
\footnotesize
\setlength{\tabcolsep}{4pt}
\begin{tabular}{lccccccc}
\toprule
Normalization, $\beta$ & Test & Circuit PR & Nearly unused & Act. & Abl. & Random & Norm-ranked \\
\midrule
None, $0$ & $0.993$ & $0.611\,(0.015)$ & $81.0$ & $0.338$ & $0.417$ & $0.669$ & $0.292$ \\
None, $1$ & $0.974$ & $0.991\,(0.003)$ & $0$ & $0.978$ & $0.916$ & $0.776$ & $0.594$ \\
Batch, $0$ & $1.000$ & $0.808\,(0.007)$ & $42.4$ & $0.735$ & $0.698$ & $0.805$ & $0.556$ \\
Batch, $1$ & $1.000$ & $0.970\,(0.003)$ & $0.2$ & $0.892$ & $0.845$ & $0.834$ & $0.651$ \\
Layer, $0$ & $1.000$ & $0.356\,(0.004)$ & $164.0$ & $0.336$ & $0.307$ & $0.584$ & $0.195$ \\
Layer, $1$ & $1.000$ & $0.369\,(0.005)$ & $161.0$ & $0.352$ & $0.328$ & $0.605$ & $0.214$ \\
\bottomrule
\end{tabular}
\end{center}
\end{table}

\subsection{A language model}
\label{app:lm}
We test whether the per-neuron condition decides which MLP units survive in a language model. The model is a GPT-2-style transformer with layer normalization before each block, six blocks, model width $384$, MLP width $1536$ with GELU, context $256$ and tied embeddings, about $30$M parameters. It is trained on TinyStories \citep{eldan2023tinystories} for $393$M tokens ($24{,}000$ steps of $64$ sequences) with Adam ($\beta_1 = 0.9$, $\beta_2 = 0.95$), learning rate $2 \cdot 10^{-3}$ with warmup and cosine decay, gradient clipping at $1.0$ and coupled L2 decay $3 \cdot 10^{-5}$ on all weight matrices. As in the multilayer perceptrons, a unit is nearly unused if its outgoing norm is below a tenth of its block's mean. The balanced condition scales the decay of each unit's incoming row and outgoing column by the ratio of that unit's gradient norm to the block mean, which leaves the mean decay unchanged and corresponds to the decay control of Appendix~\ref{app:decay} (Table~\ref{tab:lm}).

\begin{table}[!ht]
\caption{MLP units in a $30$M-parameter language model after $393$M tokens (means over seeds). Random is the random-deletion tolerance of the MLP units.}
\label{tab:lm}
\begin{center}
\footnotesize
\setlength{\tabcolsep}{4pt}
\begin{tabular}{lccccc}
\toprule
Condition & Seeds & Validation loss & Nearly unused & Circuit PR & Random \\
\midrule
Adam with L2 & $5$ & $1.378$ & $0.239$ & $0.737$ & $0.422$ \\
Adam with L2, per-unit decay balance & $5$ & $1.367$ & $0.013$ & $0.988$ & $0.489$ \\
Adam with L2, ratio held & $3$ & $1.388$ & $0.307$ & $0.670$ & --- \\
AdamW, decay $0.1$ & $3$ & $1.330$ & $0.009$ & $0.99$ & $0.487$ \\
\bottomrule
\end{tabular}
\end{center}
\end{table}

Balancing each unit's decay against its gradient removes $95\%$ of the nearly unused units and lowers validation loss by $0.007$ to $0.014$ in every seed, and random-deletion tolerance rises. Equalizing the gradient while holding each unit's ratio of gradient to decay does not restore the reference but raises the unused fraction to $0.307$, possibly because the per-unit factors are noisier under minibatches. Units die outgoing column first, as in the modular-addition networks, and most of them in the first two blocks, where about half of the units end nearly unused. AdamW reaches a lower loss with fewer than $1\%$ nearly unused units, so Adam with L2 serves here to test the condition at scale and not as a recommended setting.

\section{Theory of credit equalization}
\label{app:theory}
We use a diagonal linear network to examine how equalizing gradient magnitudes changes the distribution of contributions across units. The model allows an exact comparison between unmodified gradient flow, per-unit equalization and groupwise equalization, and it explains a mechanism for the redistribution; the ReLU-network experiments test the effect itself. We first record two elementary properties of the participation ratio and of the intervention.

\begin{proposition}[Participation ratio bounds]
\label{prop:pr}
For $a \in \R^n_{\geq 0}$, $a \neq 0$, the participation ratio of Equation~\ref{eq:pr} satisfies $1/n \leq \mathrm{PR}(a) \leq 1$. The upper bound is attained exactly when all $a_j$ are equal, and the lower bound exactly when a single $a_j$ is nonzero. Consequently $n \cdot \mathrm{PR}(a)$ interpolates between $1$ and $n$ and acts as an effective count of contributing units.
\end{proposition}
\begin{proof}
By Cauchy--Schwarz, $(\sum_j a_j)^2 \leq n \sum_j a_j^2$ with equality exactly for constant $a$, which gives $\mathrm{PR} \leq 1$. Since the $a_j$ are nonnegative, $(\sum_j a_j)^2 \geq \sum_j a_j^2$ with equality exactly when at most one entry is nonzero, which gives $\mathrm{PR} \geq 1/n$.
\end{proof}

\begin{proposition}[Credit rescaling preserves a descent direction]
\label{prop:descent}
Let $g \neq 0$ be the loss gradient of the weight matrix to which Equation~\ref{eq:knob} is applied, with one row $g_j$ per neuron of the layer it feeds, and let $\tilde{g}$ be the rescaled signal for any $\beta \in [0,1]$. Then $\langle \tilde{g}, g \rangle > 0$, so a sufficiently small step along $-\tilde{g}$ decreases the loss when the other parameters are held fixed.
\end{proposition}
\begin{proof}
Each nonzero row is rescaled by the positive scalar $s_j = \big((1-\beta)\|g_j\| + \beta \bar{g}\big) / \|g_j\|$, and zero rows remain zero, so
$\langle \tilde{g}, g \rangle = \sum_{j:\, \|g_j\| > 0} s_j \|g_j\|^2 > 0$ whenever $g \neq 0$. The directional derivative of the loss along $-\tilde{g}$ is therefore negative.
\end{proof}
This statement concerns the rescaled loss gradient before Adam and weight decay.

\begin{proposition}[Unmodified and equalized flow select different minima]
\label{prop:select}
Consider a diagonal linear network with $n$ units. Unit $i$ has incoming weight $u_i$, outgoing weight $v_i$ and contribution $a_i = u_i v_i$. Initialize $u_i = v_i = \sqrt{\epsilon_i} > 0$ with $\sum_i \epsilon_i < 1$, and train by gradient flow on the loss
\begin{equation}
L \;=\; \tfrac{1}{2}\Big(\sum\nolimits_i a_i - 1\Big)^{2},
\label{eq:toyloss}
\end{equation}
whose minima form the set $\{a : \sum_i a_i = 1\}$. In this model, equalization rescales the two-coordinate gradient $(\partial_{u_i} L, \partial_{v_i} L)$ of each unit by one common factor, so both weights of a unit are updated symmetrically and stay balanced, $u_i = v_i$ and $a_i = u_i^2$. Participation is computed from the contributions $a_i$. Unless the $\epsilon_i$ are all equal, unmodified and equalized gradient flow converge to different minima:
\begin{itemize}
\item[(i)] Unmodified gradient flow keeps every ratio $a_i / a_j$ at its initial value, so it converges to the rescaled initialization $a_i = \epsilon_i / \sum_j \epsilon_j$, whose participation ratio equals that of the initialization.
\item[(ii)] The symmetric per-unit equalized flow keeps every difference $u_i - u_j$ at its initial value, so it converges to $u_i = \sqrt{\epsilon_i} + C$ with $C$ set by $\sum_i (\sqrt{\epsilon_i} + C)^2 = 1$.
\item[(iii)] $\mathrm{PR}\big((\sqrt{\epsilon_i} + C)^2\big)$ is nondecreasing in $C \geq 0$, strictly increasing unless the $\epsilon_i$ are equal. Whenever the initialization undershoots the constraint, $\sum_i \epsilon_i < 1$ and $C > 0$, the equalized solution has strictly higher participation than the initialization, and it reaches $1$ as the initialization scale tends to zero. Overshooting initializations are outside this statement because some coordinates can cross zero under equalization.
\end{itemize}
\end{proposition}
\begin{proof}
Write $r = \sum_j a_j - 1$ for the residual, a single scalar shared by every unit. The two flows differ in one line,
\begin{equation}
\text{unmodified flow: }\ \dot{u}_i = -\partial_{u_i} L = -r\,u_i,
\qquad\qquad
\text{equalized flow: }\ \dot{u}_i = -r\,\bar{u},
\quad \bar{u} = \tfrac{1}{n}\textstyle\sum_j u_j,
\label{eq:twoflows}
\end{equation}
where the equalized form holds because unit $i$'s two-coordinate gradient $(r u_i,\, r u_i)$ has norm $|r|\sqrt{2}\,u_i$, and equalization rescales it to the mean norm $|r|\sqrt{2}\,\bar{u}$. For (i), $\dot{a}_i = 2 u_i \dot{u}_i = -2 r a_i$, so $\frac{d}{dt} \log(a_i / a_j) = 0$ because $r$ is common to all units. The invariance is a property of this model, since in a network whose units see different residuals the ratios do move. For (ii), every $u_i$ takes the same step, so $\sqrt{a_i(t)} = \sqrt{\epsilon_i} + C(t)$ with $C(t) = -\int_0^t r \bar{u}\, ds$, and fitting the target fixes $C$. For (iii), write $b_i = \sqrt{\epsilon_i} + C$ and $S_k = \sum_i b_i^k$, so that $\mathrm{PR} = S_2^2 / (n S_4)$ and
\begin{equation}
\frac{d}{dC}\,\mathrm{PR} \;=\; \frac{4\,S_2\,\big(S_1 S_4 - S_2 S_3\big)}{n\,S_4^{2}},
\qquad
S_1 S_4 - S_2 S_3 \;=\; \tfrac{1}{2}\sum_{i,j} b_i b_j\,(b_i + b_j)(b_i - b_j)^2 \;\geq\; 0,
\label{eq:prmono}
\end{equation}
where the identity follows by symmetrizing the double sum $\sum_{i,j} b_i b_j^4 - b_i^2 b_j^3$ over $i$ and $j$, and it is zero exactly when all nonzero $b_i$ coincide. Participation is nondecreasing along the path from $C = 0$, where it equals $\mathrm{PR}(\epsilon)$, to the fitted $C > 0$, and strictly increasing unless the $\epsilon_i$ are equal. As the initialization scale tends to zero, $C \to 1/\sqrt{n}$ and $a_i \to 1/n$, whose participation ratio is $1$.
\end{proof}

Under unmodified gradient flow every contribution grows by the same factor, so their ratios stay fixed. Under symmetric per-unit equalization every $u_i$ receives the same increment; the initial differences then become small relative to the shared growth, and participation increases. Both flows optimize the same loss but select different members of its set of minima. The model differs from the ReLU experiment in two ways: it rescales both weights of a unit so that they stay balanced, whereas the experiment rescales only the incoming weights, and it measures the products $u_i v_i$ rather than the outgoing norms. It provides a mechanism for the observed redistribution, not an exact prediction for the trained nonlinear network. The same model describes groupwise equalization, which preserves the ratios within each group while equalizing the group totals.

\begin{proposition}[Group equalization removes only between-group inequality]
\label{prop:grain}
Partition the $n$ units into $g$ groups of $m = n/g$, fixed throughout training. In the model of Proposition~\ref{prop:select}, rescale each unit's two-coordinate gradient by the factor, common within its group, that equalizes the total credit across groups. Write
\begin{equation}
p_i \;=\; \frac{\sqrt{\epsilon_i}}{\sum_{j \in \mathrm{group}(i)} \sqrt{\epsilon_j}}
\label{eq:groupshare}
\end{equation}
for unit $i$'s share of its own group's initialization, measured in the $\sqrt{a}$ coordinate in which the flow is linear, and let $k$ run over groups. As the initialization scale tends to zero, the fitted solution has participation ratio
\begin{equation}
\mathrm{PR}(m) \;=\; \frac{S^2}{m\,Q},
\qquad
S = \frac{1}{g}\sum_{k}\sum_{i \in k} p_i^2,
\qquad
Q = \frac{1}{g}\sum_{k}\sum_{i \in k} p_i^4,
\label{eq:grainlaw}
\end{equation}
which equals the unmodified-flow value of Proposition~\ref{prop:select} at $g=1$ and equals $1$ at $m=1$.
\end{proposition}

\begin{proof}
The rescaling factor is shared within a group, so inside group $k$ every $\log u_i$ has the same time derivative. The within-group shares $p_i$ freeze at their initial values, as the ratios freeze under unmodified flow. The group totals $U_k = \sum_{i \in k} u_i$ obey
\begin{equation}
\dot{U}_k \;=\; -\,r\,c_k U_k \;=\; -\,r\,\bar{U},
\qquad c_k = \bar{U}/U_k,
\qquad \bar{U} = \tfrac{1}{g}\textstyle\sum_k U_k,
\end{equation}
a common increment, so as the initialization tends to zero all group totals equalize to a common $U$. The solution is then $a_i = p_i^2 U^2$ with $U^2 = 1/(gS)$ forced by $\sum_i a_i = 1$, and substituting into Equation~\ref{eq:pr} gives Equation~\ref{eq:grainlaw}. At $m=1$ every share is $1$, so $S = Q = 1$ and $\mathrm{PR} = 1$, recovering Proposition~\ref{prop:select}. At $g=1$ the rescaling is the identity, and Equation~\ref{eq:grainlaw} reduces to the participation of the normalized initialization, the unmodified-flow value.
\end{proof}

Groupwise equalization changes the totals assigned to the groups while preserving the relative weights inside each group. To compare Equation~\ref{eq:grainlaw} with the neural-network sweep, we assume lognormally distributed initial contributions with log standard deviation $\sigma$. For many units the unmodified model then has participation $e^{-\sigma^2}$; we set this equal to the measured whole-layer value $0.602$, which fixes $\sigma$ from that one point. The model then predicts participation $0.674$, $0.737$, $0.834$ and $1.000$ for groups of eight, four, two and one, against measured values of $0.623$, $0.695$, $0.753$ and $0.992$. It reproduces the shape of the curve: flat for large groups and steep at the smallest.

\begin{proposition}[Participation and score fluctuations under random deletion]
\label{prop:fluct}
Model the score of the correct class as a sum of unit contributions, $y = \sum_j c_j$ with $c_j \geq 0$ over $n$ units, and delete units independently with probability $\rho$, giving the surviving score $y_\rho = \sum_j b_j c_j$ with $b_j \sim \mathrm{Bernoulli}(1-\rho)$ marking survival. Then, assuming $\sum_j c_j>0$ and $0\leq \rho<1$, for every $\varepsilon \in (0,1)$,
\begin{equation}
\frac{\sqrt{\mathrm{Var}[y_\rho]}}{\mathbb{E}[y_\rho]}
= \sqrt{\frac{\rho}{1-\rho}} \cdot \frac{1}{\sqrt{n \cdot \mathrm{PR}(c)}},
\qquad
\mathbb{P}\big(y_\rho < (1-\varepsilon)\,\mathbb{E}[y_\rho]\big)
\;\leq\; \frac{\rho}{(1-\rho)\,\varepsilon^2 \, n \cdot \mathrm{PR}(c)},
\label{eq:fluct}
\end{equation}
where $\mathrm{PR}(c)$ is Equation~\ref{eq:pr} evaluated on the vector of contributions.
\end{proposition}
\begin{proof}
$\mathbb{E}[y_\rho] = (1-\rho) \sum_j c_j$ and, by independence, $\mathrm{Var}[y_\rho] = \rho(1-\rho) \sum_j c_j^2$. Dividing and using Equation~\ref{eq:pr} gives the first identity, and Chebyshev's inequality gives the tail bound.
\end{proof}

In this additive model, a larger effective count reduces the relative fluctuation of the score under independent random deletion. The proposition gives the direction of the effect; the deletion experiments establish the accuracy differences in the trained networks.

\section{Supplementary analysis of generalization}
\label{app:generalization}

Section~\ref{sec:grokrules} reports that a weight-space measurement, $q$, separates the runs that generalize from those that only fit the training set. This appendix gives the definition, the compared configurations, the full results and the limits of the measurement.

\paragraph{A local measure on the weight vector.} A well-studied account of grokking explains generalization through weight decay: a generalizing circuit fits the training data with smaller weights than a memorizing one, so regularization gradually favors it \citep{nanda2023progress,varma2023circuit,liu2022omnigrok}. Once the training data are fitted, the network can keep moving among parameter settings that preserve the fit, and weight decay pushes it toward the smallest of them, unless some other component of the learning signal opposes it. To measure this, let $J(\theta)$ be the Jacobian of all training logits with respect to the parameters. A small parameter change $\vv$ changes the logits by $J\vv$ to first order, so the directions with $J\vv = 0$, the null space of $J$, are the directions in which weights can be reduced without changing the training outputs to first order. Let $P$ be the orthogonal projector onto this null space. Then $P\theta$ is the component of the current weight vector along these directions, and we measure its norm,
\begin{equation}
q(\theta) \;=\; \|P\theta\|_2 .
\label{eq:flatmass}
\end{equation}
Both $J$ and $P$ depend on the current parameters, so $q$ is a local measurement that we recompute at every checkpoint; it is not a count of unnecessary parameters or neurons.

For continuous dynamics $\dot{\theta} = -\nabla \ell - \lambda\theta - \vb$, where $\ell$ is the training loss, $\lambda > 0$ the weight-decay strength and $\vb$ the learning signal's departure from the loss gradient, the loss gradient has no component in the null space and
\begin{equation*}
P\dot{\theta} \;=\; -\lambda P\theta - P\vb .
\end{equation*}
Weight decay shrinks the projected weights while an additional signal can oppose it. If the Jacobian is held fixed and $P\vb = 0$, the projected weights decay exponentially and $q$ tends to zero; a constant nonzero $P\vb$ instead leaves the stationary value $P\theta = -P\vb/\lambda$ (Proposition~\ref{prop:persist}). This calculation motivates measuring $q$ during training. Since the projector changes and Adam transforms the signal, we measure $q$ for every rule, including backpropagation.

\paragraph{Compared configurations.} We measure nine configurations on modular addition at $p = 31$ with three seeds each and a $15{,}000$-epoch budget: the reference, full credit equalization, DFC, feedback alignment (FA) and direct feedback alignment (DFA) \citep{lillicrap2016feedback,nokland2016dfa}, sign-SGD \citep{bernstein2018signsgd}, and three constructed controls. Every configuration writes its learning signal into the gradient slot, and the signal is then clipped and passed to Adam at the shared learning rate $10^{-3}$ and weight decay $10^{-3}$. Sign-SGD also memorizes when applied directly, at learning rates $10^{-4}$ and $10^{-5}$ and with weight decay up to $10^{-2}$ (three seeds, test accuracy $0.000$). Each control replaces the update by the unit vector $\cos\varphi\,\hat{\vg} + \sin\varphi\,\hat{\vv}_\perp$ rescaled to the gradient norm, where $\hat{\vg}$ is the normalized cross-entropy gradient, $\hat{\vv}_\perp$ is a random vector projected orthogonally to the current gradient and normalized, and $\cos\varphi = 0.4$. \emph{Fresh noise} draws the random vector anew at every step. \emph{Fixed bias} draws it once per run, so the added component keeps its direction up to the projection. \emph{Error-gated bias} uses the same fixed vector but multiplies the orthogonal component by the current loss relative to its initial value before the renormalization, so its cosine to the gradient is $0.4$ at the start and rises toward one as the loss falls. Fresh noise and error-gated bias grok, fixed bias memorizes. The recorded checkpoints, at epochs $75$, $150$, $300$, $450$, $750$ and $1500$, show the delay in every grokking run: training accuracy is $1.0$ by epoch $150$ (by epoch $300$ for fresh noise), and test accuracy first exceeds $0.95$ between epochs $750$ and $1500$ for the reference and error-gated bias, after epoch $1500$ for credit equalization and fresh noise, and between epochs $450$ and $750$ for DFC, whose delays at $10$-epoch resolution are $520$ to $610$ epochs (Appendix~\ref{app:repro}).

\paragraph{Results.} At epoch $300$, $2\%$ of training, all $27$ runs have training accuracy $1.0$ and test accuracy $0.0$. The estimated $q$ nevertheless already separates the eventual outcomes: the largest value among the runs that later grok is $22.02$, below the smallest value among the runs that memorize, $27.71$ (Table~\ref{tab:early_flatmass}). The separation is also present at the endpoints, where every grokking run has a smaller $q$ than every memorizing run (Table~\ref{tab:flatmass}). Total parameter norm does not separate the runs at epoch $300$: the largest norm among the runs that later grok, $42.2$, exceeds the smallest norm among the memorizing runs, $41.3$. At the endpoints the norm separates the outcomes as well (grokking runs between $13.1$ and $26.7$, memorizing runs between $35.6$ and $150$). These observations agree with the weight-decay account.

\paragraph{Within one rule.} The nine configurations differ in their rule, and each one groks in every seed or memorizes in every seed, so a separation across them could reflect the rule rather than the outcome. We therefore vary one setting within two rules, with five seeds per setting and the budget, checkpoints and estimator above (Table~\ref{tab:withinrule}). For backpropagation with Adam over weight decays from $0$ to $3\times10^{-3}$, every run with weight decay groks and the runs without it do not generalize within the budget; at weight decay $10^{-2}$ the runs do not fit the training set and are excluded. $q$ at epoch $300$ is lower in every grokking run than in every run without decay, but the runs at weight decay $10^{-4}$ grok with $q = 31.3$, above the smallest value among the memorizing configurations of Table~\ref{tab:early_flatmass}. For the fixed-bias rule, raising the cosine between update and gradient from $0.4$ to $0.95$ lowers $q$ at epoch $300$ from $48.8$ to $18.6$ and raises final test accuracy from $0.00$ to $0.98$, with a Spearman correlation of $-0.97$ over the $40$ runs. Runs at cosines up to $0.8$ do not meet the grokking criterion, although test accuracy reaches $0.46$ to $0.89$ at $0.7$ and $0.8$, and at $0.85$, $0.9$ and $0.95$ two, four and four of five seeds grok. No single threshold on $q$ at epoch $300$ separates the runs that grok from the others in this family (largest value among the grokking runs $24.9$, smallest among the others $18.6$), and within a setting the differences in $q$ between seeds, about $0.1$, do not consistently order the outcomes. Within a rule, $q$ thus changes with a setting in the direction the weight-decay account predicts, but a common threshold on $q$ early in training does not predict the outcome.

\begin{table}[!ht]
\caption{Null-space weight norm $q$ within one rule (modular addition, $p=31$, five seeds per setting, $15{,}000$ epochs; standard deviations in parentheses). Grokking seeds meet the criterion of Section~\ref{sec:setup}; final test is the mean test accuracy at the end of training.}
\label{tab:withinrule}
\begin{center}
\small
\begin{tabular}{lcccc}
\toprule
Setting & $q$ at epoch $300$ & $q$ at epoch $750$ & Grokking seeds & Final test \\
\midrule
\multicolumn{5}{l}{\emph{Backpropagation with Adam, weight decay}} \\
$0$ & $35.97\,(0.13)$ & $37.33$ & $0/5$ & $0.000$ \\
$10^{-4}$ & $31.27\,(0.07)$ & $22.91$ & $5/5$ & $0.995$ \\
$3\times10^{-4}$ & $24.96\,(0.05)$ & $13.44$ & $5/5$ & $0.997$ \\
$10^{-3}$ & $14.34\,(0.08)$ & $7.57$ & $5/5$ & $0.991$ \\
$3\times10^{-3}$ & $8.38\,(0.10)$ & $4.96$ & $5/5$ & $0.996$ \\
$10^{-2}$ & $5.71\,(0.12)$ & $1.69$ & training set not fitted & $0.162$ \\
\midrule
\multicolumn{5}{l}{\emph{Fixed-bias rule, cosine between update and gradient}} \\
$0.4$ & $48.82\,(0.19)$ & $49.71$ & $0/5$ & $0.000$ \\
$0.5$ & $42.80\,(0.15)$ & $42.21$ & $0/5$ & $0.000$ \\
$0.6$ & $37.50\,(0.10)$ & $35.42$ & $0/5$ & $0.026$ \\
$0.7$ & $32.58\,(0.12)$ & $29.08$ & $0/5$ & $0.548$ \\
$0.8$ & $27.56\,(0.08)$ & $22.81$ & $0/5$ & $0.837$ \\
$0.85$ & $24.91\,(0.06)$ & $19.58$ & $2/5$ & $0.941$ \\
$0.9$ & $21.99\,(0.04)$ & $16.27$ & $4/5$ & $0.972$ \\
$0.95$ & $18.64\,(0.06)$ & $12.47$ & $4/5$ & $0.981$ \\
\bottomrule
\end{tabular}
\end{center}
\end{table}

\begin{table}[!ht]
\caption{Estimated null-space weight norm $q$ early in training, as means (standard deviations) over three seeds. Every run fits the training set and has zero test accuracy at epoch $300$. The outcome column refers to the completed runs.}
\label{tab:early_flatmass}
\begin{center}
\small
\begin{tabular}{lccc}
\toprule
Configuration & Epoch $300$ ($2\%$) & Epoch $750$ ($5\%$) & Outcome \\
\midrule
Reference (backpropagation with Adam) & $14.32\,(0.09)$ & $7.63\,(0.14)$ & groks \\
Credit equalization ($\beta{=}1$) & $14.46\,(0.08)$ & $9.11\,(0.13)$ & groks \\
DFC & $11.69\,(0.14)$ & $6.48\,(0.08)$ & groks \\
Error-gated bias & $21.93\,(0.09)$ & $9.53\,(0.04)$ & groks \\
Fresh noise & $15.37\,(0.26)$ & $9.32\,(0.13)$ & groks \\
DFA & $27.80\,(0.10)$ & $26.05\,(0.36)$ & memorizes \\
FA & $28.45\,(0.21)$ & $27.63\,(0.31)$ & memorizes \\
Fixed bias & $48.94\,(0.08)$ & $49.71\,(0.05)$ & memorizes \\
Sign-SGD & $49.13\,(0.02)$ & $72.33\,(0.48)$ & memorizes \\
\bottomrule
\end{tabular}
\end{center}
\end{table}

\begin{table}[!ht]
\caption{Estimated null-space weight norm $q$ at the end of training ($15{,}000$ epochs of modular addition at $p=31$; means over three seeds). Every grokking configuration has a smaller $q$ than every memorizing one. At initialization, $q = 29.9$.}
\label{tab:flatmass}
\begin{center}
\small
\begin{tabular}{lcllcl}
\toprule
Configuration & $q(\theta)$ & Outcome & Configuration & $q(\theta)$ & Outcome \\
\midrule
Fresh noise & $5.30$ & groks & DFA & $26.34$ & memorizes \\
Reference & $5.90$ & groks & FA & $26.55$ & memorizes \\
Error-gated bias & $6.55$ & groks & Fixed bias & $51.80$ & memorizes \\
Credit equalization ($\beta{=}1$) & $7.24$ & groks & Sign-SGD & $134.72$ & memorizes \\
DFC & $8.49$ & groks & & & \\
\bottomrule
\end{tabular}
\end{center}
\end{table}

\paragraph{Numerical estimation.} We estimate the projection as
\[
\widehat{P\theta} \;=\; \theta - J^\top (J J^\top + \epsilon I)^{-1} J\theta ,
\]
where the subtracted term approximates the component of $\theta$ in the span of the training-logit gradients and the damping $\epsilon > 0$ makes the remainder an approximation rather than an exact projection. The linear system is solved with conjugate gradients using Jacobian-vector products, without forming $J$. The reported measurements use relative damping $10^{-6}$ and $400$ conjugate-gradient iterations. On small networks the estimate agrees with an explicit singular-value decomposition to a relative error below $10^{-4}$. A longer solve at one trained width-$256$ reference checkpoint ($10{,}000$ preconditioned iterations) gives $q = 6.04$ with a relative residual of $10^{-3}$ and an idempotence error of $0.010$ relative to the parameter norm. At epoch $300$, retraining the nine configurations and re-estimating $q$ with relative damping from $10^{-10}$ to $10^{-4}$, up to $10{,}000$ iterations and in double precision changes every value by at most $0.01$, and the largest value among the runs that later grok stays below the smallest among those that memorize under every setting.

\begin{proposition}[Projected weight dynamics with a fixed local Jacobian]
\label{prop:persist}
Let $\ell$ be a differentiable loss of the training logits $z(\theta)$, let $J = \partial z / \partial \theta$, and let $P$ project orthogonally onto $\operatorname{null}(J)$. Under $\dot{\theta} = -\nabla_\theta \ell - \lambda\theta - \vb$ with $\lambda > 0$,
\begin{equation}
P\dot{\theta} \;=\; -\lambda P\theta - P\vb .
\label{eq:null_dynamics}
\end{equation}
At a stationary point, $P\theta = -P\vb/\lambda$. In the local model in which $J$ is held fixed, write $\vv = P\theta$. If $P\vb = \vb_0$ is constant, then
\[
\vv(t) \;=\; e^{-\lambda t}\vv(0) - \frac{1 - e^{-\lambda t}}{\lambda}\,\vb_0 ,
\]
so $\|\vv(t)\|_2$ tends to $\|\vb_0\|_2/\lambda$. If $P\vb(t)$ is locally integrable and tends to zero, then $\vv(t)$ tends to zero.
\end{proposition}
\begin{proof}
The chain rule gives $\nabla_\theta \ell = J^\top \nabla_z \ell$, so $P \nabla_\theta \ell = 0$. Projecting the dynamics gives Equation~\ref{eq:null_dynamics} and the stationary-point condition. For fixed $J$, the integrating-factor solution is
\[
\vv(t) \;=\; e^{-\lambda t}\vv(0) - \int_0^t e^{-\lambda (t-s)} P\vb(s)\, ds .
\]
Constant forcing gives the displayed expression. If the forcing tends to zero, splitting the integral into a fixed initial interval and the remaining tail shows that both contributions tend to zero.
\end{proof}
The proposition concerns a continuous-time model with a fixed Jacobian and the signal applied directly.

\end{document}